\documentclass[a4paper,conference]{IEEEtran}
\usepackage{pdfpages}
\usepackage{comment}
\usepackage{cite}
\usepackage{booktabs}
\usepackage{tikz}
\def\checkmark{\tikz\fill[scale=0.4](0,.35) -- (.25,0) -- (1,.7) -- (.25,.15) -- cycle;} 
\usepackage{hyperref}
\usepackage{subfigure}

\ifCLASSINFOpdf
  \usepackage{graphicx}
\else
\fi

\usepackage{amsmath}
\usepackage{amssymb}
\usepackage{relsize}
\usepackage{url}
\usepackage{color}
\usepackage{float}
\usepackage{wrapfig}

\definecolor{cadmiumgreen}{rgb}{0.0, 0.42, 0.24}
\definecolor{auburn}{rgb}{0.43, 0.21, 0.1}

\begin{document}

%
\title{Segment Augmentation and Differentiable Ranking for Logo Retrieval}

\author{\IEEEauthorblockN{Feyza Yavuz}
\IEEEauthorblockA{Dept. of Computer Engineering\\ Middle East Technical University, Ankara, Turkey \\
Email: feyza.yavuz@metu.edu.tr}
\and
\IEEEauthorblockN{Sinan Kalkan}
\IEEEauthorblockA{Dept. of Computer Engineering\\ Middle East Technical University, Ankara, Turkey  \\
Email: skalkan@metu.edu.tr}
}

\maketitle

\begin{abstract}
Logo retrieval is a challenging problem since the definition of similarity is more subjective than image retrieval, and the set of known similarities is very scarce. In this paper, to tackle this challenge, we propose a simple but effective segment-based augmentation strategy to introduce artificially similar logos for training deep networks for logo retrieval. In this novel augmentation strategy, we first find segments in a logo and apply transformations such as rotation, scaling, and color change, on the segments, unlike the conventional strategies that perform augmentation at the image level. Moreover, we evaluate suitability of using ranking-based losses (namely Smooth-AP) for learning similarity for logo retrieval. On the METU and the LLD datasets, we show that (i) our segment-based augmentation strategy improves retrieval performance compared to the baseline model or image-level augmentation strategies, and (ii) Smooth-AP indeed performs better than conventional losses for logo retrieval.
\end{abstract}


%

\IEEEpeerreviewmaketitle

\section{Introduction}

\begin{figure}[hbt!]
\centering
\includegraphics[width=0.99\columnwidth]{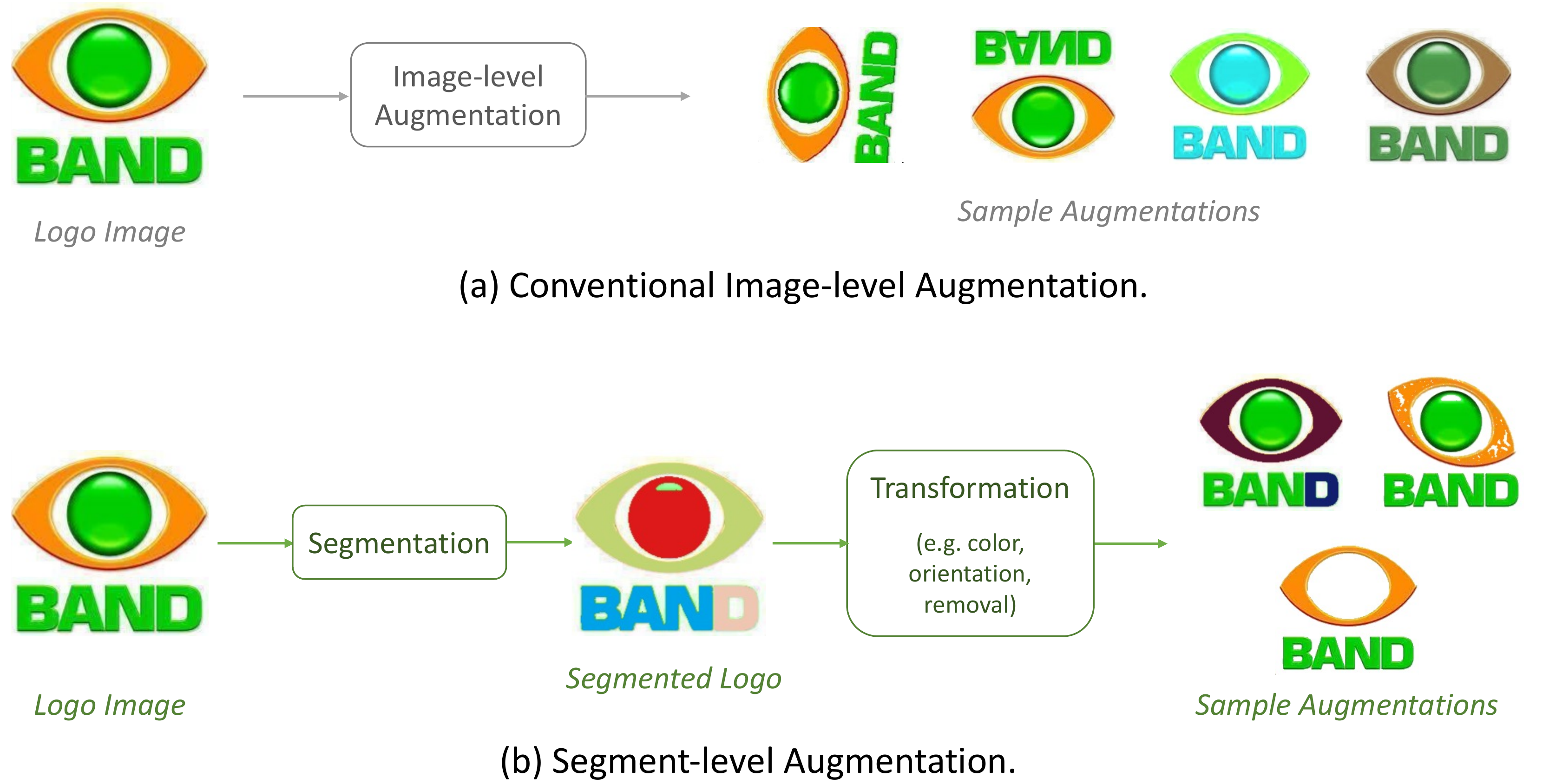}
\caption{(a) Conventional data augmentation approaches apply transformations at the image level. (b) We propose segment-level augmentation as a more suitable approach for problems like logo retrieval.
\label{fig_teaser}}
\end{figure}

With the rapid increase in companies founded worldwide and the fierce competition among them globally, the identification of companies with their logos has become more pronounced, and it has become more paramount to check similarities between logos to prevent trademark infringements. Checking trademark infringements is generally performed manually by experts, which can be sub-optimal due to human-caused errors and time-consuming as it takes days to make a decision. Therefore, automatically identifying similar logos using content-based image processing techniques is crucial.

For a query logo, identifying the similar ones in a database of logos is a content-based image retrieval problem. With the rise in deep learning, there have been many studies that have used deep learning for logo retrieval problem \cite{Tursun2019, Feng, unsupervisedattention, Tursun2022, Perez}. Existing approaches generally rely on extracting features of logos and ranking them according to a suitable distance metric \cite{TursunandKalkan, Tursun2019, Perez}.

Logo retrieval is a challenging problem especially for two main reasons: (i) Similarity between logos is highly subjective, and similarity can occur at different levels, e.g., texture, color, segments and their combination etc. (ii) The amount of known similar logos is limited. We hypothesize that this has limited the use of more modern deep learning solutions, e.g. metric learning, contrastive learning, differentiable ranking, as they require tremendous amount of positive pairs (similar logos) as well as negative pairs (dissimilar logos) for training deep networks.

In this paper, we address these challenges by (i) proposing a segment-level augmentation to produce artificially similar logos and (ii) using metric learning (Triplet Loss \cite{weinberger2009distance}) and differentiable ranking (Smooth Average Precision (AP) Loss \cite{Brown2020eccv}) as a proof of concept that, with our novel segment-augmentation method, such data hungry techniques can be trained better. 

\textit{Main Contributions.} Our contributions are as follows:
\begin{itemize}
    \item We propose a segment-level augmentation for producing artificial similarities between logos. To the best of our knowledge, ours is the first to introduce segment-level augmentation into deep learning. Unlike image-level augmentation methods that transform the overall image, we identify segments in a logo and make transformations at the segment level. Our results suggest that this is more suitable than image-level augmentation for logo retrieval.
    
    \item To showcase the use of such a tool to generate artificially similar logos, we use data-hungry deep learning methods, namely, Triplet Loss \cite{weinberger2009distance} and Smooth-AP Loss \cite{Brown2020eccv}, to show that our novel segment-augmentation method can indeed yield better retrieval performance. To the best of our knowledge, ours is the first to use such methods for logo retrieval.
\end{itemize}

\section{Related Work}

\subsection{Logo Retrieval}
Earlier studies in trademark retrieval \cite{TursunandKalkan} used hand-crafted features and deep features extracted using pre-trained networks and revealed that deep features obtained considerably better results. Perez \emph{et al.} \cite{Perez} improved the results by combining two CNNs trained on two different datasets. Later, Tursun \emph{et al.} \cite{Tursun2019} achieved impressive results by introducing different attention methods to reduce the effect of text regions, and in their most recent work \cite{Tursun2022}, they introduced different modifications and achieved state-of-the-art results. 

\subsection{Data Augmentation}
Data augmentation \cite{augmentation2, AugmentationSurvey} is an essential and well-known technique in deep learning to make networks more robust to variations in data.  Conventional augmentation methods perform geometric transformations such as zooming, flipping or cropping the entire image. Alternatively, adding noise, random erasing  or synthesizing training data \cite{ImageNet} are key approaches to improve overall model performance. Random Erasing \cite{RandomErasing} is a recently introduced method that obtains significant improvement on various recognition tasks. Although augmentation methods that focus on cutting and mixing windows \cite{CutMix,MixUp,imagemix}  rather than the whole image are not widely used, they have shown significant gains in performance. 

In logo retrieval, studies generally use conventional augmentation methods. For example, Tursun \emph{et al.} \cite{tursun2021learning} applied a reinforcement learning approach to learn an ensemble of test-time data augmentations for trademark retrieval. An exception to such an approach is the study by Tursun \emph{et al.} \cite{Tursun2019}, who proposed a method to remove text regions from logos while evaluating similarity. 

\subsection{Differentiable Ranking}

Image or logo retrieval are by definition ranking problems, though ranking is not differentiable. To address this limitation, many solutions have been proposed recently \cite{FastAP,BlackboxAP,Brown2020eccv}. These approaches mainly optimize Average Precision (AP) with different approximations: For example, Cakir \emph{et al.} \cite{FastAP} quantize distances between pairs of instances and use differentiable relaxations for these quantized distances. Rolinek \emph{et al.} \cite{BlackboxAP} consider non-differentiable ranking as a black box and use smoothing to estimate suitable gradients for training a network to rank. Finally, Brown \emph{et al.} \cite{Brown2020eccv} propose smoothing AP itself to use differentiable operations to train a deep network to rank.


These approximations have been mainly applied to standard retrieval benchmarks. In this paper, we show that differentiable ranking-based loss functions can lead to a performance improvement for logo retrieval as well.

\subsection{Summary}

Looking at the studies in the literature, we observe that \textbf{(1)} No study has performed segment-level augmentation either for logo retrieval or for general recognition or retrieval problems. The closest study for this research direction is the study by Tursun \emph{et al.} \cite{Tursun2019}, which just removed text regions in logos while evaluating similarity. \textbf{(2)} Promising deep learning approaches such as metric learning using e.g. Triplet Loss and differentiable ranking have not been employed for logo retrieval.
\section{Method}

In this section, after providing a definition for logo retrieval, we present our novel segment-based augmentation method and how we use it with deep metric learning and differentiable ranking approaches. 

\begin{wrapfigure}{r}{0.45\columnwidth}
\centering
\includegraphics[width=0.45\columnwidth]{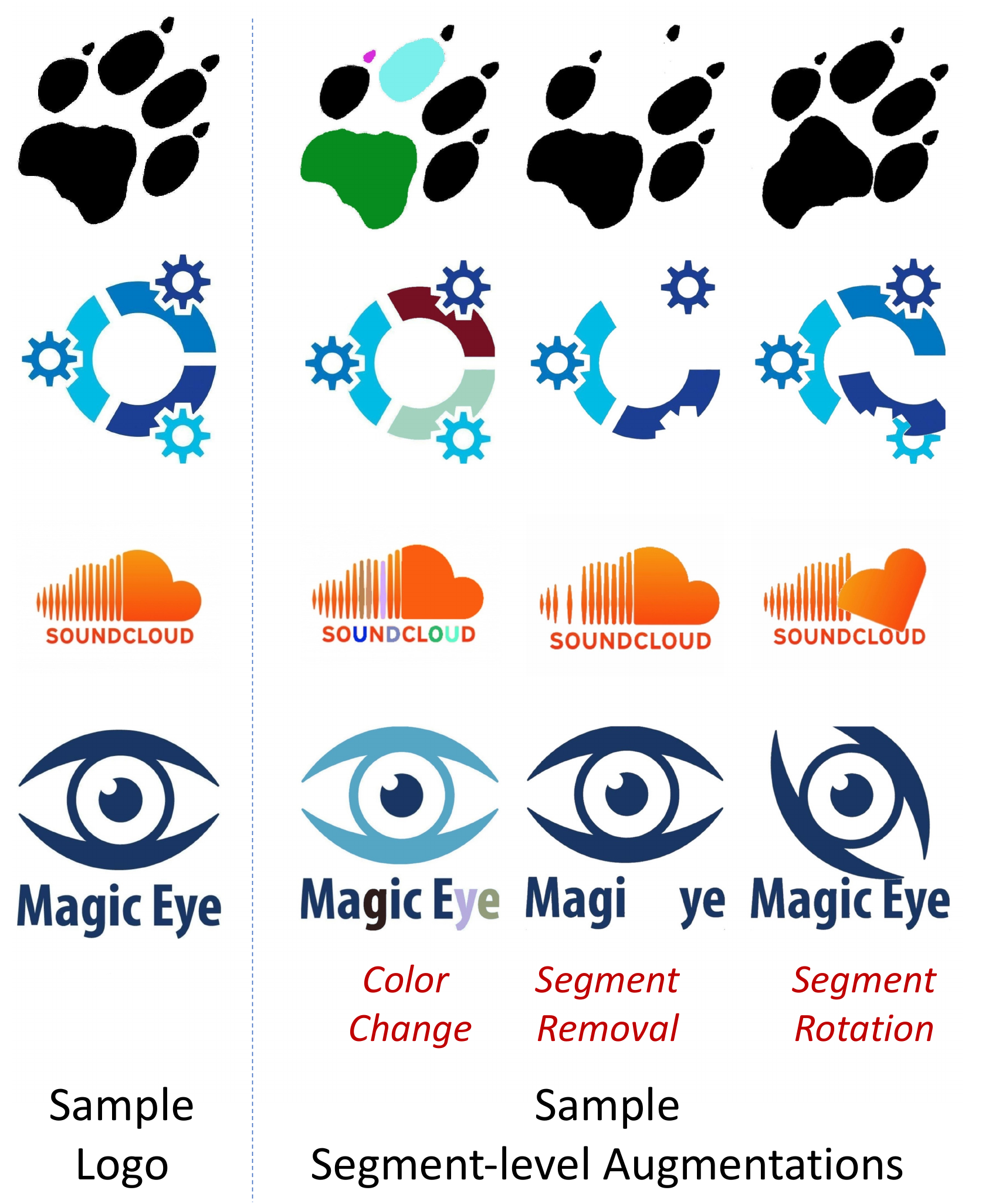}
\caption{Examples for our segment-level augmentation.}
\label{fig_augmentation}
\end{wrapfigure}

\subsection{Problem Definition}

Given an input query logo $I_q$, logo retrieval aims to rank all logos in a retrieval set $\Omega$ = ${I_i}$, $i=\{0,1,..., N\}$, based on their similarity to the query $I_q$. To be able to evaluate retrieval performance and to train a deep network that relies on known similarities, we require for each $I_q$ to have a set of positive (similar) logos, $\Omega^+(I_q)$, and a set of negative (dissimilar) logos, $\Omega^-(I_q)$. Note that logo retrieval defined as such does not have the notion of classes of a classification setting.

\subsection{Segment-level Augmentation for Logo Retrieval}

We perform segment-level augmentation by following these steps: (i) Logo segmentation, (ii) segment selection, and (iii) segment transformation. See Figure \ref{fig_augmentation} for some samples.
\begin{wrapfigure}{r}{0.45\columnwidth}
  \centering
  \vspace*{-4cm}
    \includegraphics[width=0.45\columnwidth]{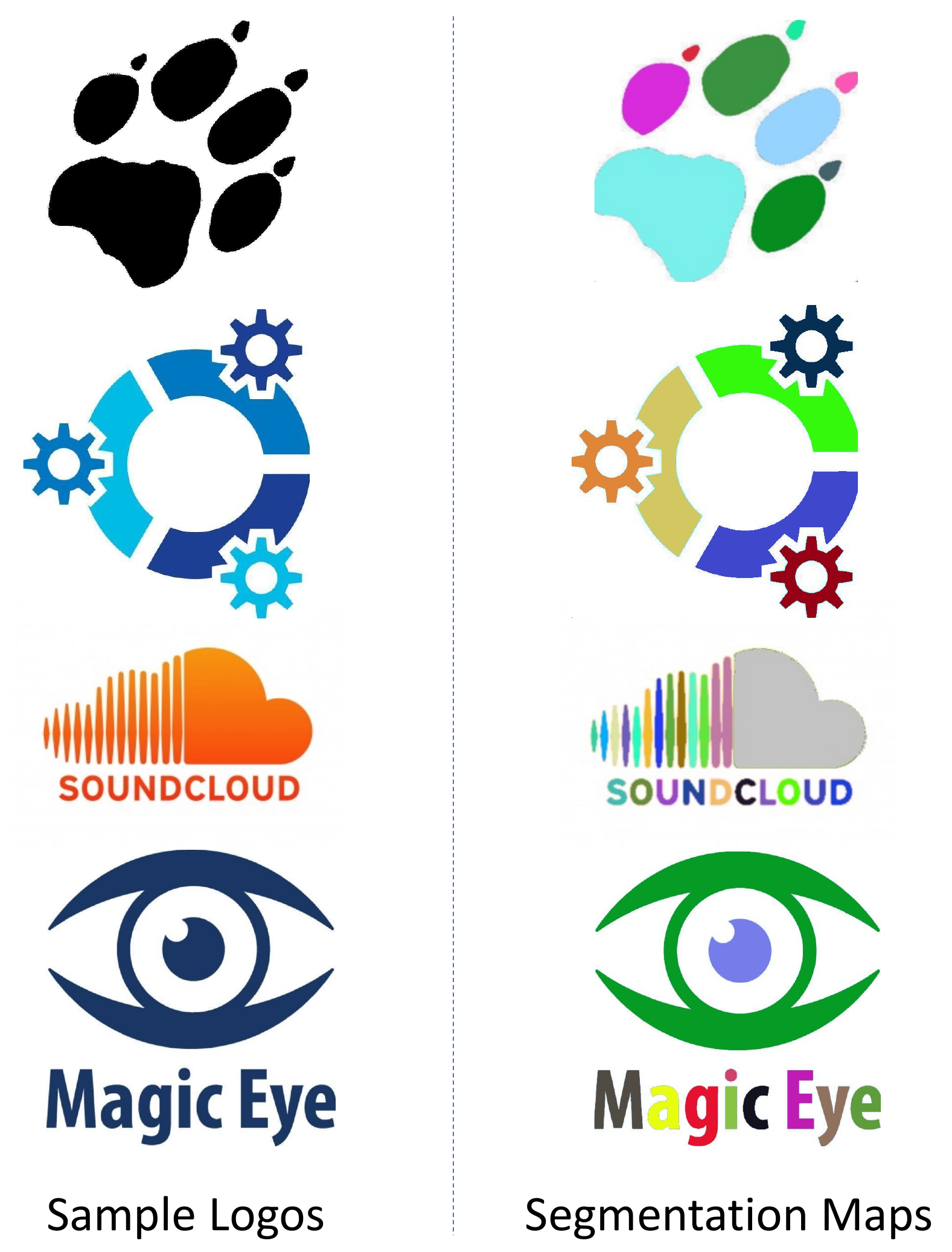}
    \caption{Sample segmentation results. Each segment is represented with different color.}
    \label{fig_segmentation}
\end{wrapfigure}


\subsubsection{Logo Segmentation}

There are many sophisticated segmentation approaches available in the literature. Since logo images have relatively simpler regions compared to images, we observed that a simple and computationally-cheap approach using standard connected-component labeling is sufficient for extracting logo segments. See Figure \ref{fig_segmentation} for some samples, and Supp. Mat. Section S4 for more samples and a discussion on the effect of segmentation quality.


\subsubsection{Segment Selection}

The next step is to select $n$ random segments to apply  transformations on them. Segment selection is a process that should be evaluated carefully since the number of segments or the area for each segment is not the same for each logo. Simplicity of logo instances also affects the number of available components and many logo instances have less than five components. Therefore, the choice of $n$ can have drastic effects especially when the number of components in a logo is small, especially for the `segment removal' transformation. For this reason, `segment removal' is not applied to a segment with the largest area, and $n$ is chosen to be small values. We present an ablation study to evaluate the effect of $n$ on model performance for the introduced augmentation strategies. For the same reason, the background component is removed from available segments for augmentation.

\subsubsection{Segment Transformation}

\begin{figure}[!h]
    \centering
    \includegraphics[width=0.9\columnwidth]{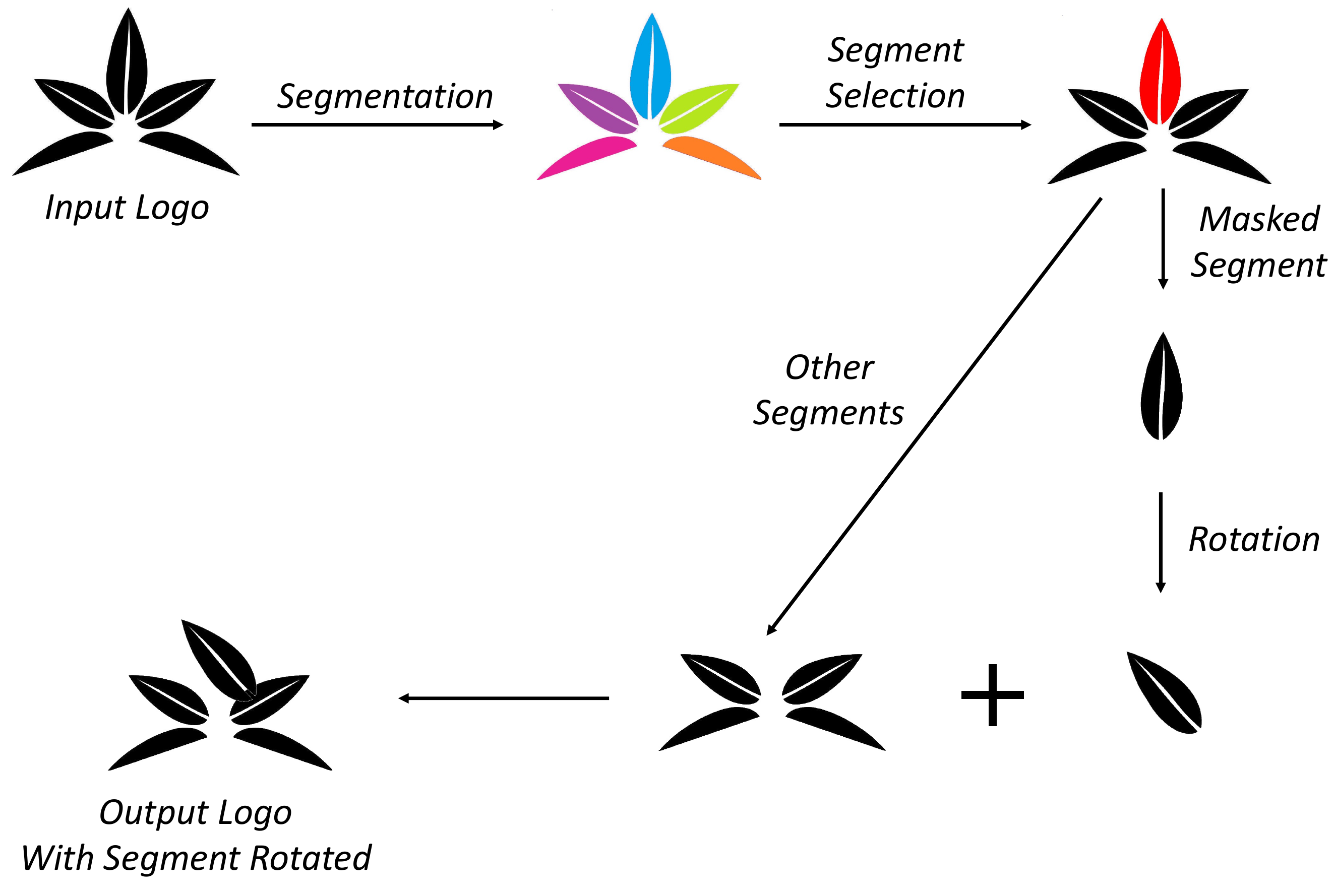}
    \caption{The steps of rotating a segment.}
    \label{fig_segment_rotation}
\end{figure}

For each selected segment $S$, the following are performed with probability $p$:
\begin{itemize}
    \item `(Segment) Color change': Every pixel in $S$ is assigned to a randomly selected color. 
    \item `Segment removal': Pixel values in $S$ are set to the same value of the background component.
    \item `Segment rotation': We first select a segment and create a mask for the segment. The mask image and the corresponding segment pixels are rotated with a random angle in [-90, 90]. Then, the rotated segment is combined with the other segments. See also Figure \ref{fig_segment_rotation} for an example.
\end{itemize}

See Figure \ref{fig_augmentation} for some sample augmentations.


\subsection{Adapting Ranking Losses for Logo Retrieval}

\subsubsection{Mini-batch Sampling for Training}
\label{sect:minibatch}
For training the deep networks, we construct the batches as follows, similar to but different from \cite{Brown2020eccv} as we do not have classes: Each mini-batch $B$ with size $|B|$ is constructed with two sub-sets: the similar set $B^{+}$ and the dissimilar set $B^{-}$. The similar logo set $B^{+}$ consists of logos that are known to be similar to each other (this information is available in the dataset \cite{Kalkan}; logos with known similarities are provided as the ``query set''), and $B^{-}$ contains logos that are dissimilar to the logos in $B^{+}$ (to be specific, logos other than the query set of the dataset are randomly sampled for $B^{-}$). The size of $B^{+}$ is set to $4$, and that of $B^{-}$ is $|B|-4$. For training the network, every $I \in B^{+}$ has label as ``1'' and $I \in B^{-}$ has label ``0''.

\subsubsection{Smooth-AP Adaptation}

Smooth-AP \cite{Brown2020eccv} is a ranking-based differentiable loss function, approximating AP. The main aspect of this approximation is to replace discrete counting operation (the indicator function) in the non-differentiable AP with a Sigmoid function. Brown et al. \cite{Brown2020eccv} applied their study to standard retrieval benchmarks such as Stanford Online Products \cite{song2016deep}, VGGFace2 \cite{DBLP:journals/corr/abs-1710-08092} and VehicleID \cite{liu2016deep}. However, the logo retrieval problem requires a dataset with a different structure as there is no notion of classes as in the Stanford Online Products \cite{song2016deep}, VGGFace2 \cite{DBLP:journals/corr/abs-1710-08092} and VehicleID \cite{liu2016deep} datasets. Hence, Smooth-AP cannot be applied directly to our problem.

The first adaptation is about the structure of the mini-batch sampling. In Smooth-AP, Brown et al. explain their sampling as they have \textit{``formed each mini-batch by randomly sampling classes such that each represented class has P samples per class''} \cite{Brown2020eccv}. Standard retrieval benchmarks have a notion of classes and are assumed to have sufficient instances per class to distribute among the mini-batches; however, there are not enough instances for known similarity ``classes'' in logo retrieval. This difference requires an adaption in both sampling and calculation of the loss. Smooth-AP Loss is calculated as follows \cite{Brown2020eccv}:
\begin{equation}
{\mathcal{L}_{AP}} = {\frac{1}{C}} {\sum_{k=1}^{C} (1-\tilde{AP}_{k})} ,
\end{equation}
where $C$ is the number of \textit{classes} and $\tilde{AP}_k$ is the smoothed AP calculated for each class in the mini-batch with their Sigmoid-based smoothing method.

Our mini-batch sampling (Section \ref{sect:minibatch}) causes a natural contradiction because our batches only contain two classes: ``similar'' and ``dissimilar''; therewith the ``dissimilar'' class should not be included in the calculation of the loss. Dissimilar class instances have the same label (``0''), but that does not mean they have the same class; they are just not similar to the similar logo set $B^+$ in the mini-batch.
Hence, the ranking among $B^-$ does not matter in our case. This difference in the batch construction and the notion of classes lead to our second adaptation. In this adaptation, the only calculated AP approximation belongs to the known ``similar'' class (logos in $B^+$). Therefore, the loss calculation becomes:
\begin{equation}
    \mathcal{L}^+_{AP} = 1-{\tilde{AP}_+},
\end{equation}
where $\tilde{AP}$ is calculated (approximated) in the same way as in the original paper \cite{Brown2020eccv}. 

\subsubsection{Triplet Loss Adaptation}

Triplet Loss \cite{weinberger2009distance} is a well-known loss function used in many computer vision problems. Triplet Loss is  differentiable, but, unlike Smooth-AP Loss \cite{Brown2020eccv}, rather than optimizing ranking, it optimizes the distances between positive pairs and negative pairs of instances. In this paper, for each mini-batch, triplets consist of one ``anchor" instance, one positive instance, and one negative instance. For the same reasons discussed in Smooth-AP Loss \cite{Brown2020eccv}, only the instances of known similarity classes can be used as the anchor instance. Optimizing the distances between dissimilar logo instances is not sensible because, as discussed in the previous section, instances of the dissimilar logos do not have any known similarity between them. Thus, triplet loss calculation is limited to the triplets that contain known similar instances as the ``anchor'' instance.
\section{Experiments and Results}

We now evaluate the performance of the proposed segment-augmentation strategy and its use with Triplet Loss and Smooth-AP Loss. 

\subsection{Experimental and Implementation Details}

\subsubsection{Dataset}

We use the METU Dataset \cite{Kalkan}, which is one of the largest publicly available logo retrieval datasets. The dataset is composed of more than 900K authentic logos belonging to actual companies worldwide. Moreover, it includes query sets, i.e. similar logos, of varying difficulties, allowing logo retrieval researchers to benchmark their methods against other methods. We have used 411K training images, 413K test images, and 418 query images. 

\subsubsection{Training and Implementation Details}

For every experiment that will be discussed, we use ImageNet \cite{ImageNet} pre-trained ResNet50 \cite{ResNet} as our backbone architecture which has a linear layer with 512 dimensions, rather than a final Softmax layer. We use the Adam optimizer with the hyper-parameters tuned as $10^{-7}$ for the learning rate and 256 for the batch size.


\subsubsection{Evaluation Measures}

Following the earlier studies \cite{Kalkan,Tursun2019}, we use  {Normalized Average Rank} (NAR) and {Recall@K} for quantifying the performance of the methods. NAR is calculated as:
\begin{equation}
  {NAR = \frac{1}{N\times N_{rel}} \left(\sum_{i=1}^{N_{rel}} R_i - \frac{N_{rel}(N_{rel}+ 1)}{2} \right)},
\end{equation}
where $N_{rel}$ is the number of similar images for a particular query image; $N$ is the size of the image set;  and $R_i$ is the rank of the $i^{th}$ similar image. NAR lies in the range $[0,1]$, where $0$ denotes the perfect score, and $1$ the worst. Recall@K (R@K) is recall for top-K similar logos.

\begin{table}[hbt!]
\centering
\caption{\label{tab:RankingLossesTable} The effect of using Triplet Loss and Smooth-AP Loss for logo retrieval.  Neither image-level nor segment-level augmentation is used for any method in this table.}
\begin{tabular}{@{}llll@{}}
\toprule
\textbf{Method}  & \textbf{NAR} $\downarrow$   & \textbf{Recall@1} $\uparrow$   & \textbf{Recall@8} $\uparrow$    \\ \midrule
Baseline       & 0.102          & 0.310          & 0.536          \\ \midrule
Triplet Loss   & 0.053          & \textbf{0.344} & \textbf{0.586} \\ \midrule
Smooth-AP Loss & \textbf{0.046} & 0.339          & 0.581          \\ \bottomrule
\addlinespace
\end{tabular}
\end{table}

\subsection{Experiment 1: Effect of Ranking Losses}

Before analyzing the effect of segment-level augmentation, in this section, we first provide a stand-alone analysis to illustrate the effect of the ranking losses. We compare Triplet Loss and Smooth-AP Loss with a baseline that compares features extracted with the pre-trained Resnet50 backbone using Cosine Similarity. For this analysis, no image-level or segment-level augmentations are used, except for the Random Resized Crop to fit the images to the expected resolution of the network, i.e.  $224 \times 224$. 

The results in Table \ref{tab:RankingLossesTable} suggest that both loss adaptations provide a significant performance improvement in both NAR and Recall measures and Smooth-AP adaptation achieves the best performance.  Applying Cosine Similarity on off-the-shelf ResNet50 features shows adequate results in no-text logo instances, however, it performs worse on logos with text (see Appendix E). 

It is evident that the improvement in Recall is not as visible as NAR. This difference states that the adapted loss functions highly affect the overall rankings of the similar known instances. However, these effects are not completely reflected by the Recall because of the selected K$=8$ value.  

\begin{table}[ht]
\centering
\caption{\label{tab:BestAugmentationResults} The effect of image-level (H. Flip, V. Flip) and segment-level augmentation. Only the best augmentation strategies are reported. See Section \ref{sect:ablation} for an ablation analysis.}
\begin{tabular}{@{}llll@{}}
\toprule
\textbf{Method}  & \textbf{NAR} $\downarrow$   & \textbf{Recall@1} $\uparrow$   & \textbf{Recall@8} $\uparrow$    \\ \midrule

Baseline\\
(No augmentation) & 0.102          & 0.310          & 0.536          \\ \midrule\midrule
Triplet Loss\\
(No augmentation)& 0.053 & 0.344 & 0.586 \\ \midrule
Triplet Loss\\
(Image-level aug.) & 0.051          & 0.354 & 0.596 \\ \midrule
Triplet Loss\\
(S. Color, S. Removal) & 0.046          & \textbf{0.374} & \textbf{0.640} \\ \midrule\midrule
Smooth-AP Loss\\
(No augmentation) & 0.046          & 0.339 & 0.581 \\ \midrule
Smooth-AP Loss\\
(Image-level aug.) & 0.044          & 0.339 & 0.596 \\ \midrule
Smooth-AP Loss \\
(S. Color)  & \textbf{0.040} & 0.354          & 0.610          \\ \bottomrule
\addlinespace
\end{tabular}
\end{table}

\begin{table}[hbt!]
\centering
\caption{\label{tab:SOTATable} Normalized Average Rank (NAR) values for previous state-of-the-art results on the METU dataset. The results are not comparable as the methods differ in their backbones, training datasets, or training regime. For some methods, these details are not even reported. See the text for details.}
\begin{tabular}{@{}ll@{}}
\toprule
\textbf{Method}  & \textbf{NAR} $\downarrow$\\ \midrule
Hand-crafted Features (Feng \emph{et al.} \cite{Feng}) & 0.083 \\ \midrule
Hand-crafted Features (Tursun \emph{et al.} \cite{Kalkan})  & 0.062 \\ \midrule
Off-the-shelf Deep Features (Tursun \emph{et al.} \cite{TursunandKalkan})  & 0.086 \\ \midrule
Transfer Learning (Perez \emph{et al.} \cite{Perez})   & 0.047  \\ \midrule
Component-based attention (SPoC \cite{SPOC}, \cite{Tursun2019}) & 0.120  \\ \midrule
Component-based attention (CRoW \cite{CRoW}, \cite{Tursun2019}) & 0.140  \\ \midrule
Component-based attention (R-MAC \cite{MAC}, \cite{Tursun2019}) & 0.072  \\ \midrule
Component-based attention (MAC \cite{MAC} \cite{Tursun2019}) & 0.120  \\ \midrule
Component-based attention (Jimenez \cite{Jimenez},\cite{Tursun2019}) & 0.093  \\ \midrule
Component-based attention (CAM MAC \cite{Tursun2019}) & 0.064  \\ \midrule
Component-based attention (ATR MAC \cite{Tursun2019}) & 0.056  \\ \midrule
Component-based attention (ATR R-MAC \cite{Tursun2019}) & 0.063  \\ \midrule
Component-based attention (ATR CAM MAC \cite{Tursun2019}) & 0.040 \\ \midrule
MR-R-MAC w/UAR (Tursun \emph{et al.} \cite{Tursun2022}) & {0.028} \\ \midrule
Segment-Augm. (Color Change) w Smooth-AP (Ours)   & {0.040}  \\
\bottomrule
\addlinespace
\end{tabular}
\end{table}

\subsection{Experiment 2: Effect of Segment Augmentation}

We now compare our segment-based augmentation methods with the conventional image-level augmentation techniques on the METU dataset \cite{Kalkan}. In every experiment, we resize the images with Random Resized Crop to fit them to the expected resolution of the network, i.e. 224$\times$224. For both segment-based and image-level augmentation, the same number of images are augmented and the probability $p=0.5$ is used for selecting a certain transformation.
 
We have provided a comparison between the best resulting methods for both image-level and segment-level augmentation methods in Table \ref{tab:BestAugmentationResults}. We see that image-level augmentation can improve ranking performance. However, the results suggest that segment-level augmentation provides a significantly better gain both in terms of NAR and R@K measures. Detailed comparison between image-level and segment-level methods is provided in ablation study.

\subsection{Experiment 3: Comparison with State of the Art}

We compare our method and the state-of-the-art methods on the METU dataset \cite{Kalkan}. It is important to note that a fair comparison between the methods is not possible because they differ in their backbones, training datasets or dataset splits, or training time. For some papers, even those details are missing; e.g. for \cite{Tursun2022}, which reports the best NAR performance. Therefore, we list the results in Table \ref{tab:SOTATable}, and refrain from drawing conclusions.

\begin{table}[ht]
    \centering
    \caption{\label{tab:ProbabilityTable} The effect of probability $p$ for augmenting a selected segment, with Smooth-AP Loss.}
    \begin{tabular}{@{}lllllll@{}}
        \toprule
        \multicolumn{1}{c}{Color C.} &
        \multicolumn{1}{c}{Rotation} &
        \multicolumn{1}{c}{Removal} &
        \multicolumn{1}{c}{\textbf{$p$}} &
        \multicolumn{1}{c}{\textbf{NAR $\downarrow$}} &
        \multicolumn{1}{c}{\textbf{R@1 $\uparrow$}} &
        \multicolumn{1}{c}{\textbf{R@8 $\uparrow$}} \\ \midrule \midrule
        \multicolumn{3}{c}{\textit{Baseline}} &
        \multicolumn{1}{c}{0} &
        \multicolumn{1}{c}{0.046} &
        \multicolumn{1}{c}{0.339} &
        \multicolumn{1}{c}{0.581}  \\ \midrule \midrule
        \multicolumn{1}{c}{\checkmark} &
        \multicolumn{1}{c}{} &
        \multicolumn{1}{c}{} &
        \multicolumn{1}{c}{0.5} &
        \multicolumn{1}{c}{\textbf{0.040}} &
        \multicolumn{1}{c}{0.354} &
        \multicolumn{1}{c}{\textbf{0.610}}  \\ \midrule
        \multicolumn{1}{c}{\checkmark} &
        \multicolumn{1}{c}{} &
        \multicolumn{1}{c}{} &
        \multicolumn{1}{c}{0.75} &
        \multicolumn{1}{c}{0.043} &
        \multicolumn{1}{c}{0.354} &
        \multicolumn{1}{c}{0.601}  \\ \midrule
        \multicolumn{1}{c}{\checkmark} &
        \multicolumn{1}{c}{} &
        \multicolumn{1}{c}{} &
        \multicolumn{1}{c}{1.0} &
        \multicolumn{1}{c}{0.049} &
        \multicolumn{1}{c}{0.325} &
        \multicolumn{1}{c}{0.566} \\ \midrule \midrule
         \multicolumn{1}{c}{} &
        \multicolumn{1}{c}{\checkmark} &
        \multicolumn{1}{c}{} &
        \multicolumn{1}{c}{0.5} &
        \multicolumn{1}{c}{0.048} &
        \multicolumn{1}{c}{0.344} &
        \multicolumn{1}{c}{0.601}  \\ \midrule
        \multicolumn{1}{c}{} &
        \multicolumn{1}{c}{\checkmark} &
        \multicolumn{1}{c}{} &
        \multicolumn{1}{c}{0.75} &
        \multicolumn{1}{c}{0.049} &
        \multicolumn{1}{c}{\textbf{0.369}} &
        \multicolumn{1}{c}{0.591}  \\ \midrule
        \multicolumn{1}{c}{} &
        \multicolumn{1}{c}{\checkmark} &
        \multicolumn{1}{c}{} &
        \multicolumn{1}{c}{1.0} &
        \multicolumn{1}{c}{0.060} &
        \multicolumn{1}{c}{0.330} &
        \multicolumn{1}{c}{0.556} \\ \midrule \midrule
        \multicolumn{1}{c}{} &
        \multicolumn{1}{c}{} &
        \multicolumn{1}{c}{\checkmark} &
        \multicolumn{1}{c}{0.5} &
        \multicolumn{1}{c}{0.048} &
        \multicolumn{1}{c}{0.339} &
        \multicolumn{1}{c}{0.596} \\ \midrule
        \multicolumn{1}{c}{} &
        \multicolumn{1}{c}{} &
        \multicolumn{1}{c}{\checkmark} &
        \multicolumn{1}{c}{0.75} &
        \multicolumn{1}{c}{0.047} &
        \multicolumn{1}{c}{0.344} &
        \multicolumn{1}{c}{0.571} \\  \midrule
        \multicolumn{1}{c}{} &
        \multicolumn{1}{c}{} &
        \multicolumn{1}{c}{\checkmark} &
        \multicolumn{1}{c}{1.0} &
        \multicolumn{1}{c}{0.047} &
        \multicolumn{1}{c}{0.344} &
        \multicolumn{1}{c}{0.591} \\
         \bottomrule
        \addlinespace
        \end{tabular}
    \end{table}

\begin{table*}[hbt!]
\centering
\caption{\label{tab:SmoothAP-DataAugmentation} Normalized Average Rank (NAR) and Recall@K values for data augmentation experiments with Smooth-AP Loss.}
 \resizebox{\textwidth}{!}{%
\begin{tabular}{@{}l l l l l | l l l l l l @{}}
\toprule
\textbf{} &
  \textbf{} &
  \textbf{Image Level} &
  \textbf{} &
   &
   & 
  \textbf{Segment Level} &
  \textbf{} &
  \textbf{} &
  \textbf{} &
   \\ \midrule
\multicolumn{1}{c}{Resized Crop} &
  \multicolumn{1}{c}{Hor. Flip} &
  \multicolumn{1}{c}{Vert. Flip} &
  \multicolumn{1}{c}{Rotation} &
  \multicolumn{1}{c}{Color Jitter} &
  \multicolumn{1}{|c}{S. Color Change} &
  \multicolumn{1}{c}{S. Rotation} &
  \multicolumn{1}{c}{S. Removal} &
  \multicolumn{1}{c}{\textbf{NAR $\downarrow$}} &
  \multicolumn{1}{c}{\textbf{R@1 $\uparrow$}} &
  \multicolumn{1}{c}{\textbf{R@8 $\uparrow$}} \\ \midrule
\multicolumn{8}{c}{\textit{Baseline}} &
  \multicolumn{1}{c}{0.102} &
  \multicolumn{1}{c}{0.310} &
  \multicolumn{1}{c}{0.536} \\ \midrule
\multicolumn{1}{c}{\checkmark} &
  \multicolumn{1}{c}{} &
  \multicolumn{1}{c}{\textbf{}} &
  \multicolumn{1}{c}{\textbf{}} &
  \multicolumn{1}{c}{} &
  \multicolumn{1}{c}{} &
  \multicolumn{1}{c}{} &
  \multicolumn{1}{c}{} &
  \multicolumn{1}{c}{0.046} &
  \multicolumn{1}{c}{0.339} &
  \multicolumn{1}{c}{0.581} \\ \midrule
\multicolumn{1}{c}{\checkmark} &
  \multicolumn{1}{c}{\checkmark} &
  \multicolumn{1}{c}{\textbf{}} &
  \multicolumn{1}{c}{\textbf{}} &
  \multicolumn{1}{c}{} &
  \multicolumn{1}{c}{} &
  \multicolumn{1}{c}{} &
  \multicolumn{1}{c}{} &
  \multicolumn{1}{c}{0.049} &
  \multicolumn{1}{c}{0.325} &
  \multicolumn{1}{c}{0.591} \\ \midrule
\multicolumn{1}{c}{\checkmark} &
  \multicolumn{1}{c}{\textbf{}} &
  \multicolumn{1}{c}{\checkmark} &
  \multicolumn{1}{c}{} &
  \multicolumn{1}{c}{} &
  \multicolumn{1}{c}{} &
  \multicolumn{1}{c}{} &
  \multicolumn{1}{c}{} &
  \multicolumn{1}{c}{0.044} &
  \multicolumn{1}{c}{0.325} &
  \multicolumn{1}{c}{0.596} \\ \midrule
\multicolumn{1}{c}{\checkmark} &
  \multicolumn{1}{c}{\checkmark} &
  \multicolumn{1}{c}{\checkmark} &
  \multicolumn{1}{c}{} &
  \multicolumn{1}{c}{} &
  \multicolumn{1}{c}{} &
  \multicolumn{1}{c}{} &
  \multicolumn{1}{c}{} &
  \multicolumn{1}{c}{0.044} &
  \multicolumn{1}{c}{0.339} &
  \multicolumn{1}{c}{0.596} \\ \midrule
\multicolumn{1}{c}{\checkmark} &
  \multicolumn{1}{c}{} &
  \multicolumn{1}{c}{} &
  \multicolumn{1}{c}{\checkmark} &
  \multicolumn{1}{c}{} &
  \multicolumn{1}{c}{} &
  \multicolumn{1}{c}{} &
  \multicolumn{1}{c}{} &
  \multicolumn{1}{c}{0.045} &
  \multicolumn{1}{c}{0.344} &
  \multicolumn{1}{c}{0.551} \\ \midrule
\multicolumn{1}{c}{\checkmark} &
  \multicolumn{1}{c}{} &
  \multicolumn{1}{c}{} &
  \multicolumn{1}{c}{} &
  \multicolumn{1}{c}{\checkmark} &
  \multicolumn{1}{c}{} &
  \multicolumn{1}{c}{} &
  \multicolumn{1}{c}{} &
  \multicolumn{1}{c}{0.049} &
  \multicolumn{1}{c}{0.349} &
  \multicolumn{1}{c}{0.576} \\ \midrule\midrule
\multicolumn{1}{c}{\checkmark} &
  \multicolumn{1}{c}{} &
  \multicolumn{1}{c}{} &
  \multicolumn{1}{c}{} &
  \multicolumn{1}{c}{} &
  \multicolumn{1}{c}{\checkmark} &
  \multicolumn{1}{c}{} &
  \multicolumn{1}{c}{} &
  \multicolumn{1}{c}{\textbf{0.040}} &
  \multicolumn{1}{c}{0.354} &
  \multicolumn{1}{c}{0.610} \\ \midrule
\multicolumn{1}{c}{\checkmark} &
  \multicolumn{1}{c}{} &
  \multicolumn{1}{c}{} &
  \multicolumn{1}{c}{} &
  \multicolumn{1}{c}{} &
  \multicolumn{1}{c}{} &
  \multicolumn{1}{c}{\checkmark} &
  \multicolumn{1}{c}{} &
  \multicolumn{1}{c}{0.048} &
  \multicolumn{1}{c}{0.344} &
  \multicolumn{1}{c}{0.601} \\ \midrule
\multicolumn{1}{c}{\checkmark} &
  \multicolumn{1}{c}{} &
  \multicolumn{1}{c}{} &
  \multicolumn{1}{c}{} &
  \multicolumn{1}{c}{} &
  \multicolumn{1}{c}{} &
  \multicolumn{1}{c}{} &
  \multicolumn{1}{c}{\checkmark} &
  \multicolumn{1}{c}{0.048} &
  \multicolumn{1}{c}{0.339} &
  \multicolumn{1}{c}{0.596} \\ \midrule
\multicolumn{1}{c}{\checkmark} &
  \multicolumn{1}{c}{} &
  \multicolumn{1}{c}{} &
  \multicolumn{1}{c}{} &
  \multicolumn{1}{c}{} &
  \multicolumn{1}{c}{\checkmark} &
  \multicolumn{1}{c}{\checkmark} &
  \multicolumn{1}{c}{} &
  \multicolumn{1}{c}{0.050} &
  \multicolumn{1}{c}{0.354} &
  \multicolumn{1}{c}{0.586} \\ \midrule
\multicolumn{1}{c}{\checkmark} &
  \multicolumn{1}{c}{} &
  \multicolumn{1}{c}{} &
  \multicolumn{1}{c}{} &
  \multicolumn{1}{c}{} &
  \multicolumn{1}{c}{\checkmark} &
  \multicolumn{1}{c}{} &
  \multicolumn{1}{c}{\checkmark} &
  \multicolumn{1}{c}{0.046} &
  \multicolumn{1}{c}{\textbf{0.374}} &
  \multicolumn{1}{c}{\textbf{0.625}} \\ \midrule
\multicolumn{1}{c}{\checkmark} &
  \multicolumn{1}{c}{} &
  \multicolumn{1}{c}{} &
  \multicolumn{1}{c}{} &
  \multicolumn{1}{c}{} &
  \multicolumn{1}{c}{} &
  \multicolumn{1}{c}{\checkmark} &
  \multicolumn{1}{c}{\checkmark} &
  \multicolumn{1}{c}{0.044} &
  \multicolumn{1}{c}{0.354} &
  \multicolumn{1}{c}{0.591} \\ \midrule
\multicolumn{1}{c}{\checkmark} &
  \multicolumn{1}{c}{} &
  \multicolumn{1}{c}{} &
  \multicolumn{1}{c}{} &
  \multicolumn{1}{c}{} &
  \multicolumn{1}{c}{\checkmark} &
  \multicolumn{1}{c}{\checkmark} &
  \multicolumn{1}{c}{\checkmark} &
  \multicolumn{1}{c}{0.047} &
  \multicolumn{1}{c}{0.374} &
  \multicolumn{1}{c}{0.605} \\ \bottomrule
  \addlinespace
\end{tabular}%
 }
\end{table*}

\begin{table*}[hbt!]
\centering
\caption{\label{tab:TripletLoss-DataAugmentation} Normalized Average Rank (NAR) and Recall@K values for data augmentation experiments with Triplet Loss.}
\resizebox{\textwidth}{!}
{%
\begin{tabular}{@{} l l l l l | l l l l l l @{}}
\toprule
\textbf{} &
  \textbf{} &
  \textbf{Image-Level} &
  \textbf{} &
   &
   &
  \textbf{Segment-Level} &
  \textbf{} &
  \textbf{} &
  \textbf{} &
   \\ \midrule
\multicolumn{1}{c}{Resized Crop} &
  \multicolumn{1}{c}{Hor. Flip} &
  \multicolumn{1}{c}{Vert. Flip} &
  \multicolumn{1}{c}{Rotation} &
  \multicolumn{1}{c}{Color Jitter} &
  \multicolumn{1}{|c}{S. Color Change} &
  \multicolumn{1}{c}{S. Rotation} &
  \multicolumn{1}{c}{S. Removal} &
  \multicolumn{1}{c}{\textbf{NAR $\downarrow$}} &
  \multicolumn{1}{c}{\textbf{R@1 $\uparrow$}} &
  \multicolumn{1}{c}{\textbf{R@8 $\uparrow$}} \\ \midrule
\multicolumn{8}{c}{\textit{Baseline}} &
  \multicolumn{1}{c}{0.102} &
  \multicolumn{1}{c}{0.310} &
  \multicolumn{1}{c}{0.536} \\ \midrule
\multicolumn{1}{c}{\checkmark} &
  \multicolumn{1}{c}{} &
  \multicolumn{1}{c}{\textbf{}} &
  \multicolumn{1}{c}{\textbf{}} &
  \multicolumn{1}{c}{} &
  \multicolumn{1}{c}{} &
  \multicolumn{1}{c}{} &
  \multicolumn{1}{c}{} &
  \multicolumn{1}{c}{0.053} &
  \multicolumn{1}{c}{0.339} &
  \multicolumn{1}{c}{0.586} \\ \midrule
\multicolumn{1}{c}{\checkmark} &
  \multicolumn{1}{c}{\checkmark} &
  \multicolumn{1}{c}{\textbf{}} &
  \multicolumn{1}{c}{\textbf{}} &
  \multicolumn{1}{c}{} &
  \multicolumn{1}{c}{} &
  \multicolumn{1}{c}{} &
  \multicolumn{1}{c}{} &
  \multicolumn{1}{c}{0.052} &
  \multicolumn{1}{c}{0.349} &
  \multicolumn{1}{c}{0.586} \\ \midrule
\multicolumn{1}{c}{\checkmark} &
  \multicolumn{1}{c}{\textbf{}} &
  \multicolumn{1}{c}{\checkmark} &
  \multicolumn{1}{c}{} &
  \multicolumn{1}{c}{} &
  \multicolumn{1}{c}{} &
  \multicolumn{1}{c}{} &
  \multicolumn{1}{c}{} &
  \multicolumn{1}{c}{0.052} &
  \multicolumn{1}{c}{0.364} &
  \multicolumn{1}{c}{0.586} \\ \midrule
\multicolumn{1}{c}{\checkmark} &
  \multicolumn{1}{c}{\checkmark} &
  \multicolumn{1}{c}{\checkmark} &
  \multicolumn{1}{c}{} &
  \multicolumn{1}{c}{} &
  \multicolumn{1}{c}{} &
  \multicolumn{1}{c}{} &
  \multicolumn{1}{c}{} &
  \multicolumn{1}{c}{0.051} &
  \multicolumn{1}{c}{0.354} &
  \multicolumn{1}{c}{0.596} \\ \midrule
\multicolumn{1}{c}{\checkmark} &
  \multicolumn{1}{c}{} &
  \multicolumn{1}{c}{} &
  \multicolumn{1}{c}{\checkmark} &
  \multicolumn{1}{c}{} &
  \multicolumn{1}{c}{} &
  \multicolumn{1}{c}{} &
  \multicolumn{1}{c}{} &
  \multicolumn{1}{c}{0.054} &
  \multicolumn{1}{c}{0.320} &
  \multicolumn{1}{c}{0.546} \\ \midrule
\multicolumn{1}{c}{\checkmark} &
  \multicolumn{1}{c}{} &
  \multicolumn{1}{c}{} &
  \multicolumn{1}{c}{} &
  \multicolumn{1}{c}{\checkmark} &
  \multicolumn{1}{c}{} &
  \multicolumn{1}{c}{} &
  \multicolumn{1}{c}{} &
  \multicolumn{1}{c}{0.053} &
  \multicolumn{1}{c}{0.344} &
  \multicolumn{1}{c}{0.591} \\ \midrule \midrule
\multicolumn{1}{c}{\checkmark} &
  \multicolumn{1}{c}{} &
  \multicolumn{1}{c}{} &
  \multicolumn{1}{c}{} &
  \multicolumn{1}{c}{} &
  \multicolumn{1}{c}{\checkmark} &
  \multicolumn{1}{c}{} &
  \multicolumn{1}{c}{} &
  \multicolumn{1}{c}{0.048} &
  \multicolumn{1}{c}{0.369} &
  \multicolumn{1}{c}{0.601} \\ \midrule
\multicolumn{1}{c}{\checkmark} &
  \multicolumn{1}{c}{} &
  \multicolumn{1}{c}{} &
  \multicolumn{1}{c}{} &
  \multicolumn{1}{c}{} &
  \multicolumn{1}{c}{} &
  \multicolumn{1}{c}{\checkmark} &
  \multicolumn{1}{c}{} &
  \multicolumn{1}{c}{0.052} &
  \multicolumn{1}{c}{0.354} &
  \multicolumn{1}{c}{0.596} \\ \midrule
\multicolumn{1}{c}{\checkmark} &
  \multicolumn{1}{c}{} &
  \multicolumn{1}{c}{} &
  \multicolumn{1}{c}{} &
  \multicolumn{1}{c}{} &
  \multicolumn{1}{c}{} &
  \multicolumn{1}{c}{} &
  \multicolumn{1}{c}{\checkmark} &
  \multicolumn{1}{c}{0.052} &
  \multicolumn{1}{c}{0.354} &
  \multicolumn{1}{c}{0.596} \\ \midrule
\multicolumn{1}{c}{\checkmark} &
  \multicolumn{1}{c}{} &
  \multicolumn{1}{c}{} &
  \multicolumn{1}{c}{} &
  \multicolumn{1}{c}{} &
  \multicolumn{1}{c}{\checkmark} &
  \multicolumn{1}{c}{\checkmark} &
  \multicolumn{1}{c}{} &
  \multicolumn{1}{c}{0.050} &
  \multicolumn{1}{c}{0.354} &
  \multicolumn{1}{c}{0.586} \\ \midrule
\multicolumn{1}{c}{\checkmark} &
  \multicolumn{1}{c}{} &
  \multicolumn{1}{c}{} &
  \multicolumn{1}{c}{} &
  \multicolumn{1}{c}{} &
  \multicolumn{1}{c}{\checkmark} &
  \multicolumn{1}{c}{} &
  \multicolumn{1}{c}{\checkmark} &
  \multicolumn{1}{c}{\textbf{0.046}} &
  \multicolumn{1}{c}{\textbf{0.374}} &
  \multicolumn{1}{c}{\textbf{0.640}} \\ \midrule
\multicolumn{1}{c}{\checkmark} &
  \multicolumn{1}{c}{} &
  \multicolumn{1}{c}{} &
  \multicolumn{1}{c}{} &
  \multicolumn{1}{c}{} &
  \multicolumn{1}{c}{} &
  \multicolumn{1}{c}{\checkmark} &
  \multicolumn{1}{c}{\checkmark} &
  \multicolumn{1}{c}{0.048} &
  \multicolumn{1}{c}{0.354} &
  \multicolumn{1}{c}{0.571} \\ \midrule
\multicolumn{1}{c}{\checkmark} &
  \multicolumn{1}{c}{} &
  \multicolumn{1}{c}{} &
  \multicolumn{1}{c}{} &
  \multicolumn{1}{c}{} &
  \multicolumn{1}{c}{\checkmark} &
  \multicolumn{1}{c}{\checkmark} &
  \multicolumn{1}{c}{\checkmark} &
  \multicolumn{1}{c}{\textbf{0.046}} &
  \multicolumn{1}{c}{0.369} &
  \multicolumn{1}{c}{0.581} \\ \bottomrule
  \addlinespace
\end{tabular}
}
\end{table*}

\subsection{Experiment 4: Ablation Study}
\label{sect:ablation}

\subsubsection{Choice of Hyper-Parameters}

Our segment-level augmentation has two hyper-parameters: The number of segments, $n$, selected for augmentation, and the probability, $p$, of applying a selected augmentation. Table \ref{tab:ProbabilityTable} shows that the best performance is obtained with $p$ as 0.5. A similar analysis for $n$ (with values 1, 2, $L/3$ and $L/2$ where $L$ is the number of segments in a logo) provided the best performance for $n$ as $L/3$. 

\subsubsection{Effects of Individual Augmentation Methods}

Tables \ref{tab:SmoothAP-DataAugmentation} (Smooth-AP Loss) and \ref{tab:TripletLoss-DataAugmentation} (Triplet Loss) list the effects of both image-level and segment-level augmentation. The tables show that, among the segment-level augmentation methods, (Segment) `Color Change' outperforms the others for both loss functions. With Triplet Loss adaptation, (Segment) `Removal' and `Rotation' provide slightly better NAR values than the baseline. Another point worth mentioning is that combining (Segment) `Rotation' or `Removal' degrades the NAR performance measure whereas the combination of (Segment) `Removal' and `Color Change' yields the best result at $Recall@8$.

\subsubsection{Experiments with a Different Backbone}

Section A in the Appendix provides an analysis using ConvNeXt \cite{liu2022convnet}, a recent, fast and strong backbone competing with transformer-based architectures. Our results without any hyper-parameter tuning are comparable to the baseline or better than the baseline with the R@8 measure.

\subsubsection{Experiments with a Different Dataset}

Section B in the Appendix reports results on the LLD dataset \cite{sage2017logodataset} that confirm our analysis on the METU dataset: We observe that segment-level augmentation provides significant gains for all measures.


\subsection{Experiment 5: Visual Results}

Section F in the Appendix provides sample retrieval results for several query logos for the baseline as well the adaptations of Triplet Loss and Smooth-AP Loss with our segment-level augmentation methods. The visual results also confirm that segment augmentation with our Smooth-AP adaptation performs best.

\section{Conclusion}

We introduced a novel data augmentation method based on image segments for training neural networks for logo retrieval. We performed segment-level augmentation by identifying segments in a logo and do transformations on selected segments. Experiments were conducted on the METU \cite{Kalkan} and LLD \cite{sage2017logodataset} datasets with ResNet \cite{ResNet} and ConvNeXt \cite{liu2022convnet} backbones and suggest significant improvements on two evaluation measures of ranking performance.  Moreover, we use metric learning and differentiable ranking with the proposed segment-augmentation method to demonstrate that our method can lead to a further boost in ranking performance. 

We note that our segment-level augmentation strategy generates similarities between logos that are rather simplistic: It is based on the assumption that two similar logos differ from each other in terms of certain segments having differences in color, orientation and presence. An important research direction is exploring more sophisticated augmentation strategies for introducing artificial similarities. However, our results suggest that even such a simplistic strategy can improve the retrieval performance significantly and therefore, our study can be considered as a first step towards developing better segment/part-level augmentation strategies.



%
\clearpage
\bibliographystyle{./IEEEtran}
\bibliography{./references}
\clearpage

\appendix

\subsection{Experiments with ConvNeXt, a Stronger Backbone}

In Table \ref{tab:RebuttalConvNextTable}, we evaluate our novel segment-level augmentation strategy using ConvNeXt \cite{liu2022convnet}, a stronger, recent, fast backbone that can compete with transformer-based architectures. Although ConvNeXt provides significantly better performance than the ResNet baseline, Table \ref{tab:RebuttalConvNextTable} confirms our experiments with ResNet that segment-level augmentation can improve logo retrieval performance. 

\begin{table}[ht]
    \centering
    \caption{\label{tab:RebuttalConvNextTable} Experiments on the ConvNeXt architecture \cite{liu2022convnet} using Smooth-AP Loss. }
    \begin{tabular}{@{}lllllll@{}}
        \toprule
        \multicolumn{1}{c}{Color C.} &
        \multicolumn{1}{c}{Rotation} &
        \multicolumn{1}{c}{Removal} &
        \multicolumn{1}{c}{\textbf{$p$}} &
        \multicolumn{1}{c}{\textbf{NAR $\downarrow$}} &
        \multicolumn{1}{c}{\textbf{R@1 $\uparrow$}} &
        \multicolumn{1}{c}{\textbf{R@8 $\uparrow$}} \\ \midrule \midrule
        \multicolumn{3}{c}{\textit{Baseline}} &
        \multicolumn{1}{c}{0} &
        \multicolumn{1}{c}{0.039} &
        \multicolumn{1}{c}{\textbf{0.438}} &
        \multicolumn{1}{c}{0.704}  \\ \midrule \midrule
        \multicolumn{1}{c}{\checkmark} &
        \multicolumn{1}{c}{} &
        \multicolumn{1}{c}{} &
        \multicolumn{1}{c}{0.5} &
        \multicolumn{1}{c}{\textbf{0.039}} &
        \multicolumn{1}{c}{0.413} &
        \multicolumn{1}{c}{\textbf{0.709}}  \\ \midrule
         \multicolumn{1}{c}{} &
        \multicolumn{1}{c}{\checkmark} &
        \multicolumn{1}{c}{} &
        \multicolumn{1}{c}{0.5} &
        \multicolumn{1}{c}{0.041} &
        \multicolumn{1}{c}{0.413} &
        \multicolumn{1}{c}{0.694}  \\ \midrule
        \multicolumn{1}{c}{} &
        \multicolumn{1}{c}{} &
        \multicolumn{1}{c}{\checkmark} &
        \multicolumn{1}{c}{0.5} &
        \multicolumn{1}{c}{0.040} &
        \multicolumn{1}{c}{0.403} &
        \multicolumn{1}{c}{0.699} \\ \midrule
        \multicolumn{1}{c}{\checkmark} &
        \multicolumn{1}{c}{\checkmark} &
        \multicolumn{1}{c}{} &
        \multicolumn{1}{c}{0.5} &
        \multicolumn{1}{c}{0.046} &
        \multicolumn{1}{c}{0.379} &
        \multicolumn{1}{c}{0.661} \\ \midrule
        \multicolumn{1}{c}{\checkmark} &
        \multicolumn{1}{c}{} &
        \multicolumn{1}{c}{\checkmark} &
        \multicolumn{1}{c}{0.5} &
        \multicolumn{1}{c}{0.045} &
        \multicolumn{1}{c}{0.374} &
        \multicolumn{1}{c}{0.684} \\ \midrule
        \multicolumn{1}{c}{} &
        \multicolumn{1}{c}{\checkmark} &
        \multicolumn{1}{c}{\checkmark} &
        \multicolumn{1}{c}{0.5} &
        \multicolumn{1}{c}{0.051} &
        \multicolumn{1}{c}{0.369} &
        \multicolumn{1}{c}{0.655} \\ \midrule
        \multicolumn{1}{c}{\checkmark} &
        \multicolumn{1}{c}{\checkmark} &
        \multicolumn{1}{c}{\checkmark} &
        \multicolumn{1}{c}{0.5} &
        \multicolumn{1}{c}{0.055} &
        \multicolumn{1}{c}{0.354} &
        \multicolumn{1}{c}{0.615} \\ \midrule
         \bottomrule
        \addlinespace
        \end{tabular}
    \end{table}

\subsection{Experiments with a Different Dataset}

Without tuning, we repeat our experiments on using a different logo retrieval dataset, namely the Large Logo Dataset (LLD) \cite{sage2017logodataset}. LLD has {61380} logos for training and {61540} logos for testing. The LLD dataset does not have a query set for providing known similarities and therefore, for the experiments on LLD, we use the query set of the METU dataset to find similarities in the LLD dataset. 

The results in Table \ref{tab:LLDtable} show that segment-level augmentation performs better than image-level augmentation in R@1 measure and on par in terms of NAR and R@8 measures. Considering that LLD is a smaller dataset than METU, we believe that segment-level augmentation can provide a larger margin in performance if tuned.

\begin{table*}[hbt!]
\centering
\caption{\label{tab:LLDtable} Normalized Average Rank (NAR) and Recall@K values for data augmentation experiments from Albumentations library with ResNet architecture on LLD Dataset.}
 \resizebox{\textwidth}{!}{%
\begin{tabular}{@{}l l l l l | l l l l l l @{}}
\toprule
\textbf{} &
  \textbf{} &
  \textbf{Image Level} &
  \textbf{} &
   &
   & 
  \textbf{Segment Level} &
  \textbf{} &
  \textbf{} &
  \textbf{} &
   \\ \midrule
\multicolumn{1}{c}{Resized Crop} &
  \multicolumn{1}{c}{Random Rotation} &
  \multicolumn{1}{c}{R. Horizontal Flip} &
  \multicolumn{1}{c}{Color Jitter} &
  \multicolumn{1}{|c}{S. Color Change} &
  \multicolumn{1}{c}{S. Rotation} &
  \multicolumn{1}{c}{S. Removal} &
  \multicolumn{1}{c}{\textbf{NAR $\downarrow$}} &
  \multicolumn{1}{c}{\textbf{R@1 $\uparrow$}} &
  \multicolumn{1}{c}{\textbf{R@8 $\uparrow$}} \\ \midrule
\multicolumn{1}{c}{\checkmark} &
  \multicolumn{1}{c}{} &
  \multicolumn{1}{c}{\textbf{}} &
  \multicolumn{1}{c}{\textbf{}} &
  \multicolumn{1}{c}{} &
  \multicolumn{1}{c}{} &
  \multicolumn{1}{c}{} &
  \multicolumn{1}{c}{0.044} &
  \multicolumn{1}{c}{0.556} &
  \multicolumn{1}{c}{0.724} \\ \midrule
\multicolumn{1}{c}{\checkmark} &
  \multicolumn{1}{c}{\checkmark} &
  \multicolumn{1}{c}{\textbf{}} &
  \multicolumn{1}{c}{\textbf{}} &
  \multicolumn{1}{c}{} &
  \multicolumn{1}{c}{} &
  \multicolumn{1}{c}{} &
  \multicolumn{1}{c}{0.068} &
  \multicolumn{1}{c}{0.527} &
  \multicolumn{1}{c}{0.679} \\ \midrule
\multicolumn{1}{c}{\checkmark} &
  \multicolumn{1}{c}{} &
  \multicolumn{1}{c}{\checkmark} &
  \multicolumn{1}{c}{\textbf{}} &
  \multicolumn{1}{c}{} &
  \multicolumn{1}{c}{} &
  \multicolumn{1}{c}{} &
  \multicolumn{1}{c}{\textbf{0.038}} &
  \multicolumn{1}{c}{0.551} &
  \multicolumn{1}{c}{\textbf{0.738}} \\ \midrule
\multicolumn{1}{c}{\checkmark} &
  \multicolumn{1}{c}{\textbf{}} &
  \multicolumn{1}{c}{} &
  \multicolumn{1}{c}{\checkmark} &
  \multicolumn{1}{c}{} &
  \multicolumn{1}{c}{} &
  \multicolumn{1}{c}{} &
  \multicolumn{1}{c}{0.039} &
  \multicolumn{1}{c}{0.551} &
  \multicolumn{1}{c}{0.719} \\ \midrule\midrule
\multicolumn{1}{c}{\checkmark} &
  \multicolumn{1}{c}{} &
  \multicolumn{1}{c}{} &
  \multicolumn{1}{c}{} &
  \multicolumn{1}{c}{\checkmark} &
  \multicolumn{1}{c}{} &
  \multicolumn{1}{c}{} &
  \multicolumn{1}{c}{0.039} &
  \multicolumn{1}{c}{\textbf{0.581}} &
  \multicolumn{1}{c}{\textbf{0.738}} \\ \midrule
\multicolumn{1}{c}{\checkmark} &
  \multicolumn{1}{c}{} &
  \multicolumn{1}{c}{} &
  \multicolumn{1}{c}{} &
  \multicolumn{1}{c}{} &
  \multicolumn{1}{c}{\checkmark} &
  \multicolumn{1}{c}{} &
  \multicolumn{1}{c}{0.058} &
  \multicolumn{1}{c}{0.527} &
  \multicolumn{1}{c}{0.687} \\ \midrule
\multicolumn{1}{c}{\checkmark} &
  \multicolumn{1}{c}{} &
  \multicolumn{1}{c}{} &
  \multicolumn{1}{c}{} &
  \multicolumn{1}{c}{} &
  \multicolumn{1}{c}{} &
  \multicolumn{1}{c}{\checkmark} &
  \multicolumn{1}{c}{0.051} &
  \multicolumn{1}{c}{0.532} &
  \multicolumn{1}{c}{0.704} \\ \midrule
\multicolumn{1}{c}{\checkmark} &
  \multicolumn{1}{c}{} &
  \multicolumn{1}{c}{} &
  \multicolumn{1}{c}{} &
  \multicolumn{1}{c}{\checkmark} &
  \multicolumn{1}{c}{\checkmark} &
  \multicolumn{1}{c}{} &
  \multicolumn{1}{c}{0.054} &
  \multicolumn{1}{c}{0.561} &
  \multicolumn{1}{c}{0.719} \\ \midrule
\multicolumn{1}{c}{\checkmark} &
  \multicolumn{1}{c}{} &
  \multicolumn{1}{c}{} &
  \multicolumn{1}{c}{} &
  \multicolumn{1}{c}{\checkmark} &
  \multicolumn{1}{c}{} &
  \multicolumn{1}{c}{\checkmark} &
  \multicolumn{1}{c}{0.056} &
  \multicolumn{1}{c}{0.517} &
  \multicolumn{1}{c}{0.704} \\ \midrule
\multicolumn{1}{c}{\checkmark} &
  \multicolumn{1}{c}{} &
  \multicolumn{1}{c}{} &
  \multicolumn{1}{c}{} &
  \multicolumn{1}{c}{} &
  \multicolumn{1}{c}{\checkmark} &
  \multicolumn{1}{c}{\checkmark} &
  \multicolumn{1}{c}{0.068} &
  \multicolumn{1}{c}{0.507} &
  \multicolumn{1}{c}{0.684} \\ \midrule
\multicolumn{1}{c}{\checkmark} &
  \multicolumn{1}{c}{} &
  \multicolumn{1}{c}{} &
  \multicolumn{1}{c}{} &
  \multicolumn{1}{c}{\checkmark} &
  \multicolumn{1}{c}{\checkmark} &
  \multicolumn{1}{c}{\checkmark} &
  \multicolumn{1}{c}{0.064} &
  \multicolumn{1}{c}{0.536} &
  \multicolumn{1}{c}{0.684} \\ \bottomrule
  \addlinespace
\end{tabular}%
 }
\end{table*}

\subsection{Comparisons with Different Image-level Augmentations}

We now compare segment-level augmentation with more image-level augmentation methods. We have selected most commonly used image-level augmentation methods ({Cutout, Elastic Transform and Channel Shuffle}) from \cite{albumentations_paper}. For this experiment, we use the same selection probability and setup as in Section IV.C. of the main paper. The results in Table \ref{tab:albumentations} show that, \textbf{without any tuning}, segment-level augmentation performs better than this new set of image-level augmentations in terms of NAR and R@8 measures whereas is inferior in terms of R@1 measure. With tuning, segment-level augmentation has potential to perform better.

\begin{table*}[hbt!]
\centering
\caption{\label{tab:albumentations} Normalized Average Rank (NAR) and Recall@K values for data augmentation experiments from Albumentations Library with Smooth-AP Loss on ResNet.}
 \resizebox{\textwidth}{!}{%
\begin{tabular}{@{}l l l l l | l l l l l l @{}}
\toprule
\textbf{} &
  \textbf{} &
  \textbf{Image Level} &
  \textbf{} &
   &
   & 
  \textbf{Segment Level} &
  \textbf{} &
  \textbf{} &
  \textbf{} &
   \\ \midrule
\multicolumn{1}{c}{Resized Crop} &
  \multicolumn{1}{c}{Cutout} &
  \multicolumn{1}{c}{Elastic Transform} &
  \multicolumn{1}{c}{Channel Shuffle} &
  \multicolumn{1}{|c}{S. Color Change} &
  \multicolumn{1}{c}{S. Rotation} &
  \multicolumn{1}{c}{S. Removal} &
  \multicolumn{1}{c}{\textbf{NAR $\downarrow$}} &
  \multicolumn{1}{c}{\textbf{R@1 $\uparrow$}} &
  \multicolumn{1}{c}{\textbf{R@8 $\uparrow$}} \\ \midrule
  \multicolumn{7}{c}{\textit{Baseline}} &
  \multicolumn{1}{c}{0.102} &
  \multicolumn{1}{c}{0.310} &
  \multicolumn{1}{c}{0.536} \\ \midrule
\multicolumn{1}{c}{\checkmark} &
  \multicolumn{1}{c}{\checkmark} &
  \multicolumn{1}{c}{\textbf{}} &
  \multicolumn{1}{c}{\textbf{}} &
  \multicolumn{1}{c}{} &
  \multicolumn{1}{c}{} &
  \multicolumn{1}{c}{} &
  \multicolumn{1}{c}{0.052} &
  \multicolumn{1}{c}{0.325} &
  \multicolumn{1}{c}{0.581} \\ \midrule
\multicolumn{1}{c}{\checkmark} &
  \multicolumn{1}{c}{} &
  \multicolumn{1}{c}{\checkmark} &
  \multicolumn{1}{c}{\textbf{}} &
  \multicolumn{1}{c}{} &
  \multicolumn{1}{c}{} &
  \multicolumn{1}{c}{} &
  \multicolumn{1}{c}{0.060} &
  \multicolumn{1}{c}{\textbf{0.379}} &
  \multicolumn{1}{c}{0.596} \\ \midrule
\multicolumn{1}{c}{\checkmark} &
  \multicolumn{1}{c}{\textbf{}} &
  \multicolumn{1}{c}{} &
  \multicolumn{1}{c}{\checkmark} &
  \multicolumn{1}{c}{} &
  \multicolumn{1}{c}{} &
  \multicolumn{1}{c}{} &
  \multicolumn{1}{c}{0.0451} &
  \multicolumn{1}{c}{0.344} &
  \multicolumn{1}{c}{0.596} \\ \midrule\midrule
\multicolumn{1}{c}{\checkmark} &
  \multicolumn{1}{c}{} &
  \multicolumn{1}{c}{} &
  \multicolumn{1}{c}{} &
  \multicolumn{1}{c}{\checkmark} &
  \multicolumn{1}{c}{} &
  \multicolumn{1}{c}{} &
  \multicolumn{1}{c}{\textbf{0.040}} &
  \multicolumn{1}{c}{0.354} &
  \multicolumn{1}{c}{0.610} \\ \midrule
\multicolumn{1}{c}{\checkmark} &
  \multicolumn{1}{c}{} &
  \multicolumn{1}{c}{} &
  \multicolumn{1}{c}{} &
  \multicolumn{1}{c}{} &
  \multicolumn{1}{c}{\checkmark} &
  \multicolumn{1}{c}{} &
  \multicolumn{1}{c}{0.048} &
  \multicolumn{1}{c}{0.344} &
  \multicolumn{1}{c}{0.601} \\ \midrule
\multicolumn{1}{c}{\checkmark} &
  \multicolumn{1}{c}{} &
  \multicolumn{1}{c}{} &
  \multicolumn{1}{c}{} &
  \multicolumn{1}{c}{} &
  \multicolumn{1}{c}{} &
  \multicolumn{1}{c}{\checkmark} &
  \multicolumn{1}{c}{0.048} &
  \multicolumn{1}{c}{0.339} &
  \multicolumn{1}{c}{0.596} \\ \midrule
\multicolumn{1}{c}{\checkmark} &
  \multicolumn{1}{c}{} &
  \multicolumn{1}{c}{} &
  \multicolumn{1}{c}{} &
  \multicolumn{1}{c}{\checkmark} &
  \multicolumn{1}{c}{\checkmark} &
  \multicolumn{1}{c}{} &
  \multicolumn{1}{c}{0.050} &
  \multicolumn{1}{c}{0.354} &
  \multicolumn{1}{c}{0.586} \\ \midrule
\multicolumn{1}{c}{\checkmark} &
  \multicolumn{1}{c}{} &
  \multicolumn{1}{c}{} &
  \multicolumn{1}{c}{} &
  \multicolumn{1}{c}{\checkmark} &
  \multicolumn{1}{c}{} &
  \multicolumn{1}{c}{\checkmark} &
  \multicolumn{1}{c}{0.046} &
  \multicolumn{1}{c}{{0.374}} &
  \multicolumn{1}{c}{\textbf{0.625}} \\ \midrule
\multicolumn{1}{c}{\checkmark} &
  \multicolumn{1}{c}{} &
  \multicolumn{1}{c}{} &
  \multicolumn{1}{c}{} &
  \multicolumn{1}{c}{} &
  \multicolumn{1}{c}{\checkmark} &
  \multicolumn{1}{c}{\checkmark} &
  \multicolumn{1}{c}{0.044} &
  \multicolumn{1}{c}{0.354} &
  \multicolumn{1}{c}{0.591} \\ \midrule
\multicolumn{1}{c}{\checkmark} &
  \multicolumn{1}{c}{} &
  \multicolumn{1}{c}{} &
  \multicolumn{1}{c}{} &
  \multicolumn{1}{c}{\checkmark} &
  \multicolumn{1}{c}{\checkmark} &
  \multicolumn{1}{c}{\checkmark} &
  \multicolumn{1}{c}{0.047} &
  \multicolumn{1}{c}{0.374} &
  \multicolumn{1}{c}{0.605} \\ \bottomrule
  \addlinespace
\end{tabular}%
 }
\end{table*}

\subsection{An Analysis of Segmentation Maps}

\begin{figure}[!h]
    \centering
    \includegraphics[width=0.9\columnwidth]{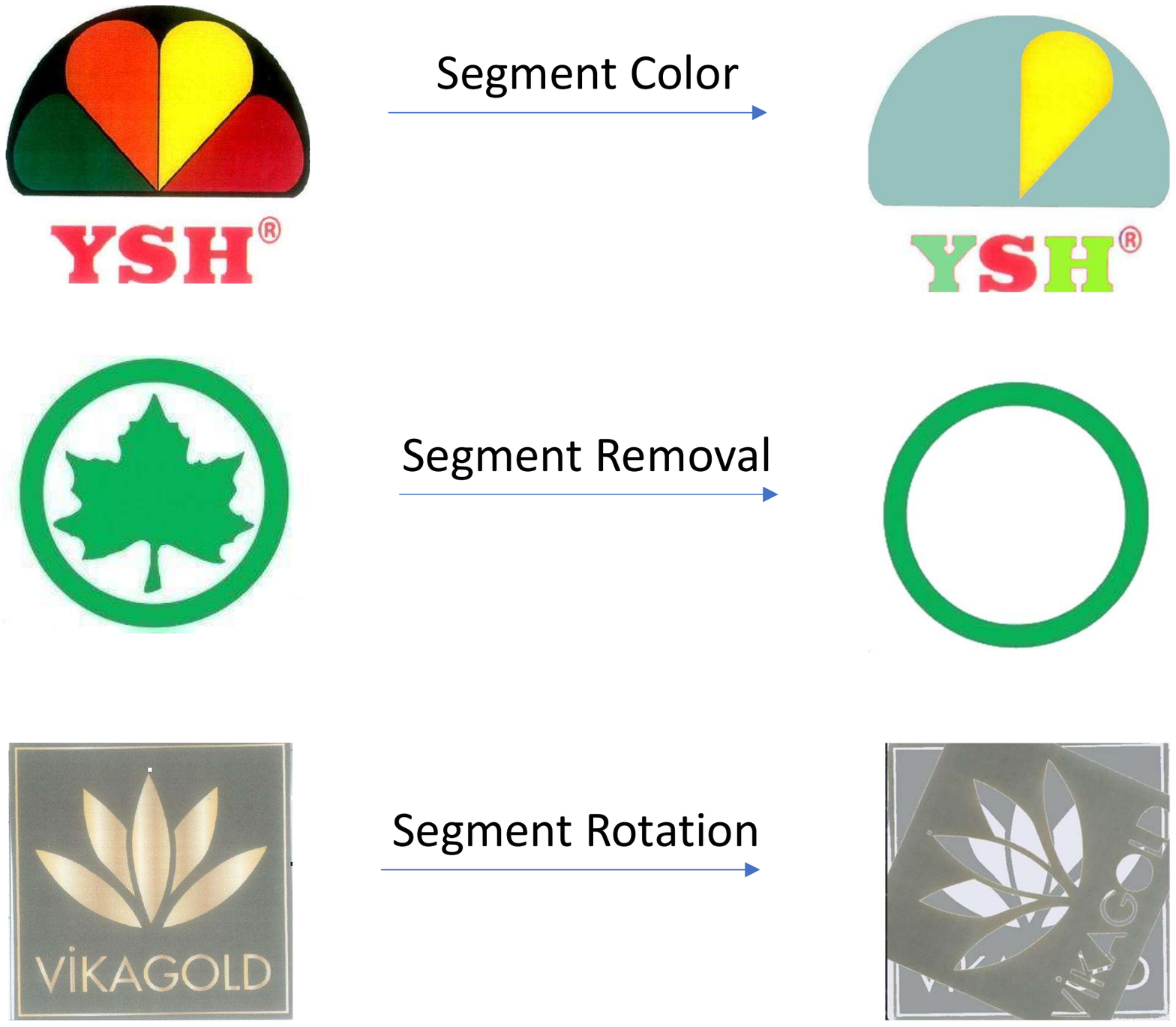}
    \caption{Edge cases in our segment-level augmentation. The first is a result of under-segmentation where the connected petals are with similar color are grouped in a single segment, and therefore, changing its color produces a distinct logo. The second case happens because the logo has two main segments and removing one results in a drastically different logo. The third is a case where a background segment is rotated and the outcome is an overlay of two segments.}
    \label{seg_fig_failure_cases}
\end{figure}

Logos are simplistic images composed of regions that are generally homogeneous in color. Therefore, an off-the-shelf simple segmentation algorithm works generally well for most logos (see Figure \ref{seg_fig_failure_cases} for some examples). Logo segmentation can produce spurious segments especially on regions which have strong color gradients. However, this is not a problem for us because segments corresponding to over-segmentation or under-segmentation do function as a form of segment-level augmentation and this is still useful for training.

Figure \ref{seg_fig_failure_cases} displays edge examples where our segment-level augmentation produces drastically different logos for which computing similarity with the original logos is highly challenging. These cases happen if there are only a few segments in a logo with comparable size and one of them is removed, or if the segmentation method under-segments the image and a segment that groups multiple regions is removed, or if the background segment is selected for augmentation and rotated. We did not try to address these edge cases as they are not frequent. We believe that they are useful stochastic perturbations and helpful to the training dynamics. We leave an analysis of this and improving the quality of segmentation \& augmentation as future work.

\subsection{The Effect of Text in Logos}

In this experiment, we analyze the effect of text on logo similarity. The visual results in Table \ref{tab:RebuttalLogoAnalysis} and the quantitative analysis in Table \ref{tab:LogoTypeAnalysis} show that, when used as queries, logos with text are more difficult to match to the logos in the dataset and the best and average ranks of similar logos are very high. This is not surprising since representations in deep networks do capture text as well and unless additional mechanisms are used to ignore them, they are taken into account while measuring similarity. The effect of text on logo retrieval is already known in the literature and soft or hard mechanisms can be used to remove them from logos -- see e.g. \cite{tursun2020learning,Kalkan}.

\begin{table*}[ht]
    \centering
    \caption{\label{tab:RebuttalLogoAnalysis} Effect of text on logo retrieval performance. When used as queries, logos with text are more difficult to match to the logos in the dataset and the best and average ranks of similar logos are very low.}
    \resizebox{\textwidth}{!}{%
    \begin{tabular}{@{}lllllllll@{}}
        \toprule
        \multicolumn{1}{c|}{\textbf{Query}} &
        \multicolumn{1}{c}{\includegraphics[width=10mm, height=10mm]{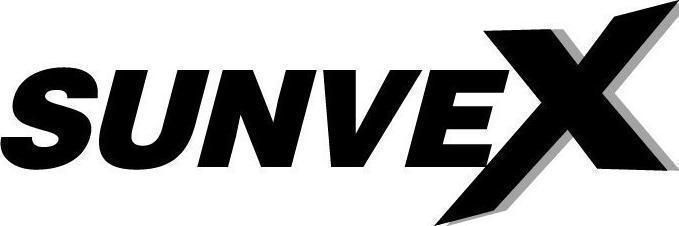}} &
        \multicolumn{1}{c|}{\includegraphics[width=10mm, height=10mm]{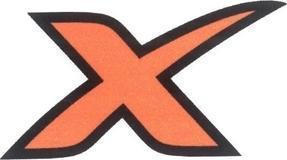}} &
        \multicolumn{1}{c}{\includegraphics[width=10mm, height=10mm]{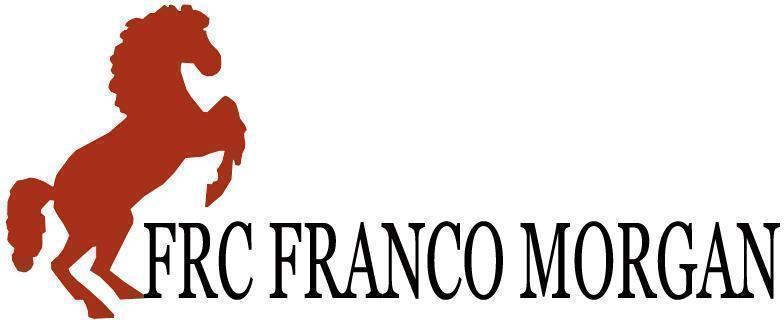}} &
        \multicolumn{1}{c|}{\includegraphics[width=10mm, height=10mm]{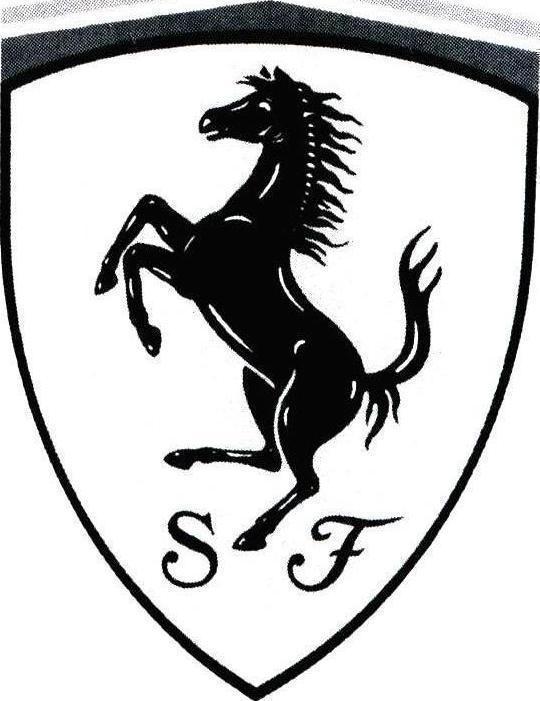}} &
        \multicolumn{1}{c}{\includegraphics[width=15mm, height=10mm]{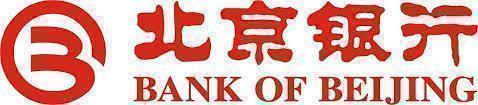}} &
        \multicolumn{1}{c|}{\includegraphics[width=10mm, height=10mm]{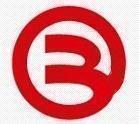}} &
        \multicolumn{1}{c}{\includegraphics[width=18mm, height=8mm]{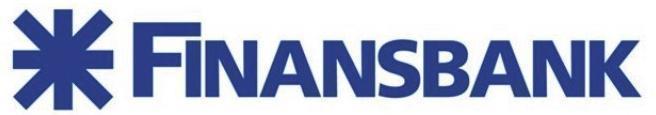}} &
        \multicolumn{1}{c}{\includegraphics[width=10mm, height=10mm]{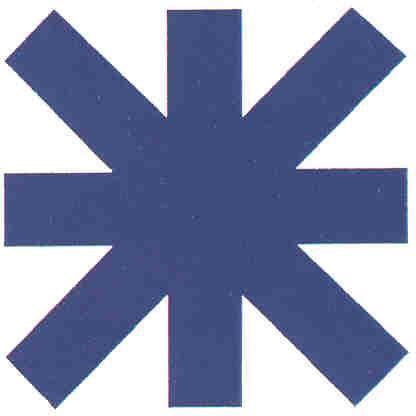}} \\ \midrule \midrule
        \multicolumn{1}{c|}{\textbf{Best Rank}} &
        \multicolumn{1}{c}{39k} &
        \multicolumn{1}{c|}{2} &
        \multicolumn{1}{c}{25} &
        \multicolumn{1}{c|}{5} &
        \multicolumn{1}{c}{97k} &
        \multicolumn{1}{c|}{1} &
        \multicolumn{1}{c}{159k} &
        \multicolumn{1}{c}{1} \\ \midrule \midrule
        \multicolumn{1}{c|}{\textbf{Average Rank}} &
        \multicolumn{1}{c}{253k} &
        \multicolumn{1}{c|}{2605} &
        \multicolumn{1}{c}{53k} &
        \multicolumn{1}{c|}{10k} &
        \multicolumn{1}{c}{239k} &
        \multicolumn{1}{c|}{25k} &
        \multicolumn{1}{c}{239k} &
        \multicolumn{1}{c}{1263} \\ 
         \bottomrule
        \addlinespace
        \end{tabular}
            }
\end{table*}

\begin{table}[ht]
\centering
\caption{\label{tab:LogoTypeAnalysis} Effect of text on logo retrieval performance using the 213 query logos in the METU Trademark Dataset.}
\begin{tabular}{@{}lll@{}}
    \toprule
    \multicolumn{1}{c}{\textbf{Logo Type}} &
    \multicolumn{1}{c}{\textbf{Number of Logos}} &
    \multicolumn{1}{c}{\textbf{NAR}} \\ \midrule \midrule
    \multicolumn{1}{c}{Logos w/o text} &
    \multicolumn{1}{c}{117} &
    \multicolumn{1}{c}{0.015} \\  \midrule
    \multicolumn{1}{c}{Logos w text} &
    \multicolumn{1}{c}{86} &
    \multicolumn{1}{c}{0.071} \\
     \bottomrule
    \addlinespace
    \end{tabular}
    \end{table}

\subsection{Visual Retrieval Results}

\begin{figure*}[!h]
\centerline{
    \subfigure[]{
        \includegraphics[width=0.65\textwidth]{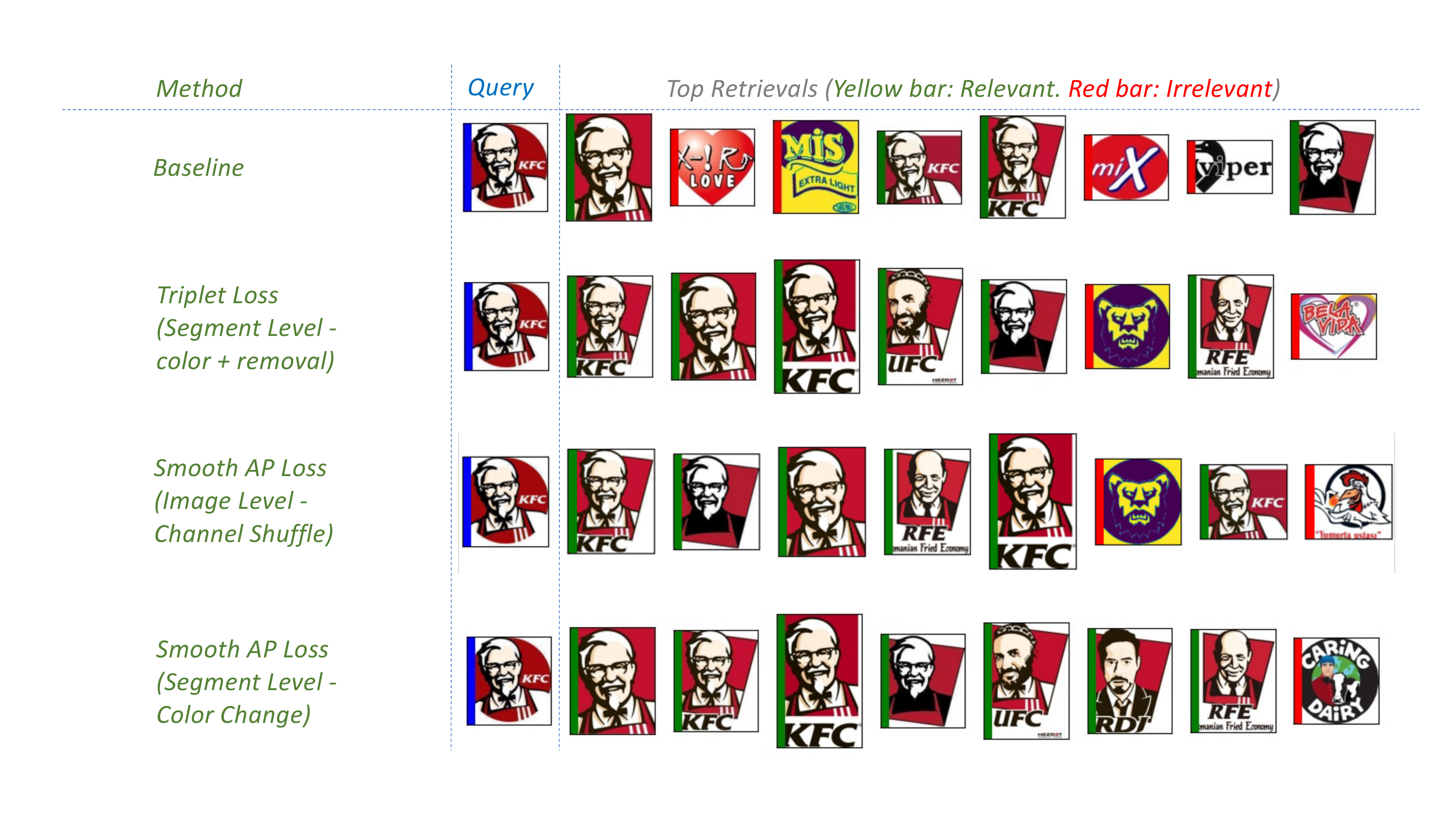}
}}
\centerline{
    \subfigure[]{
    \includegraphics[width=0.65\textwidth]{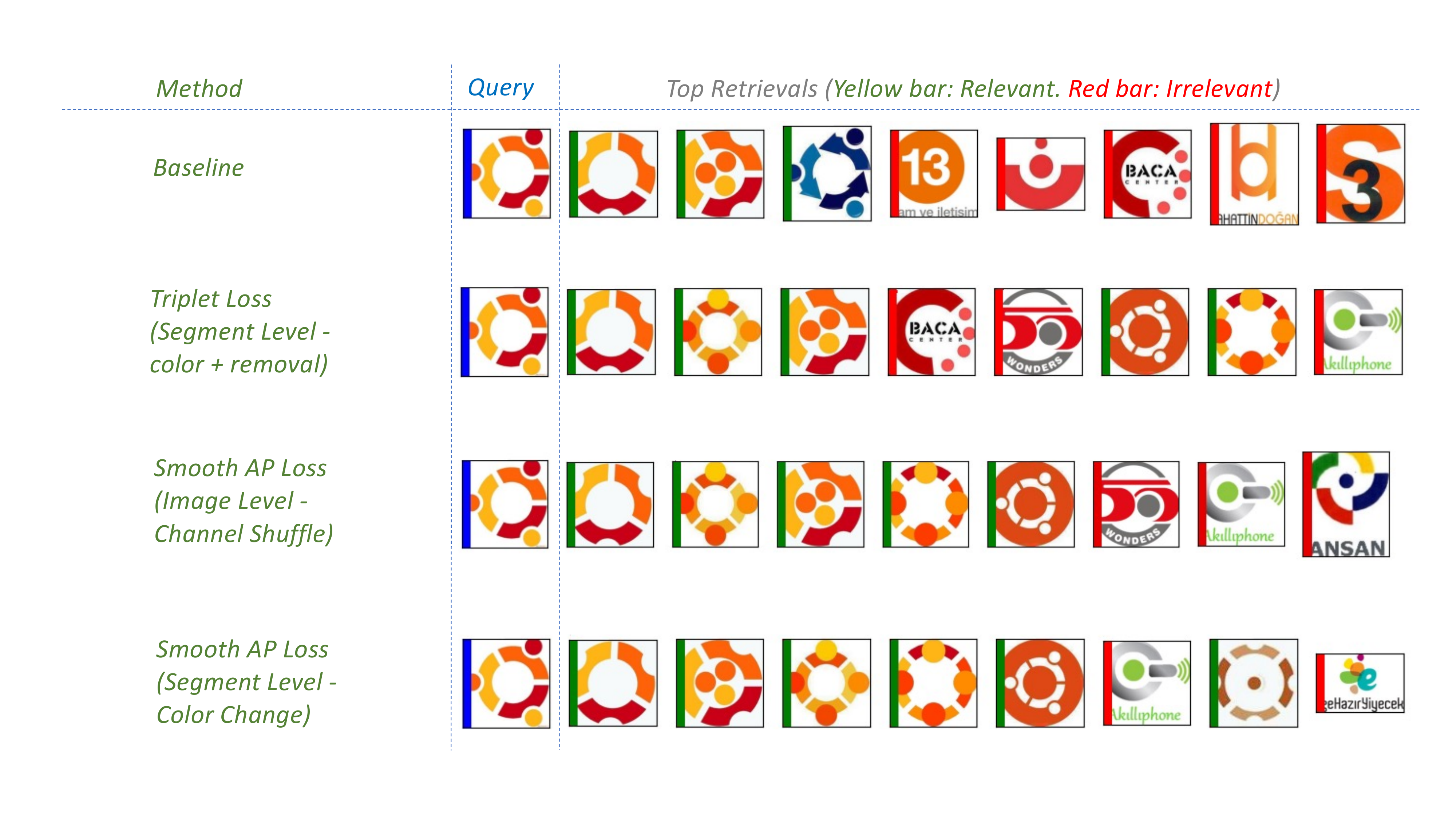}
}}
\centerline{
    \subfigure[]{
    \includegraphics[width=0.65\textwidth]{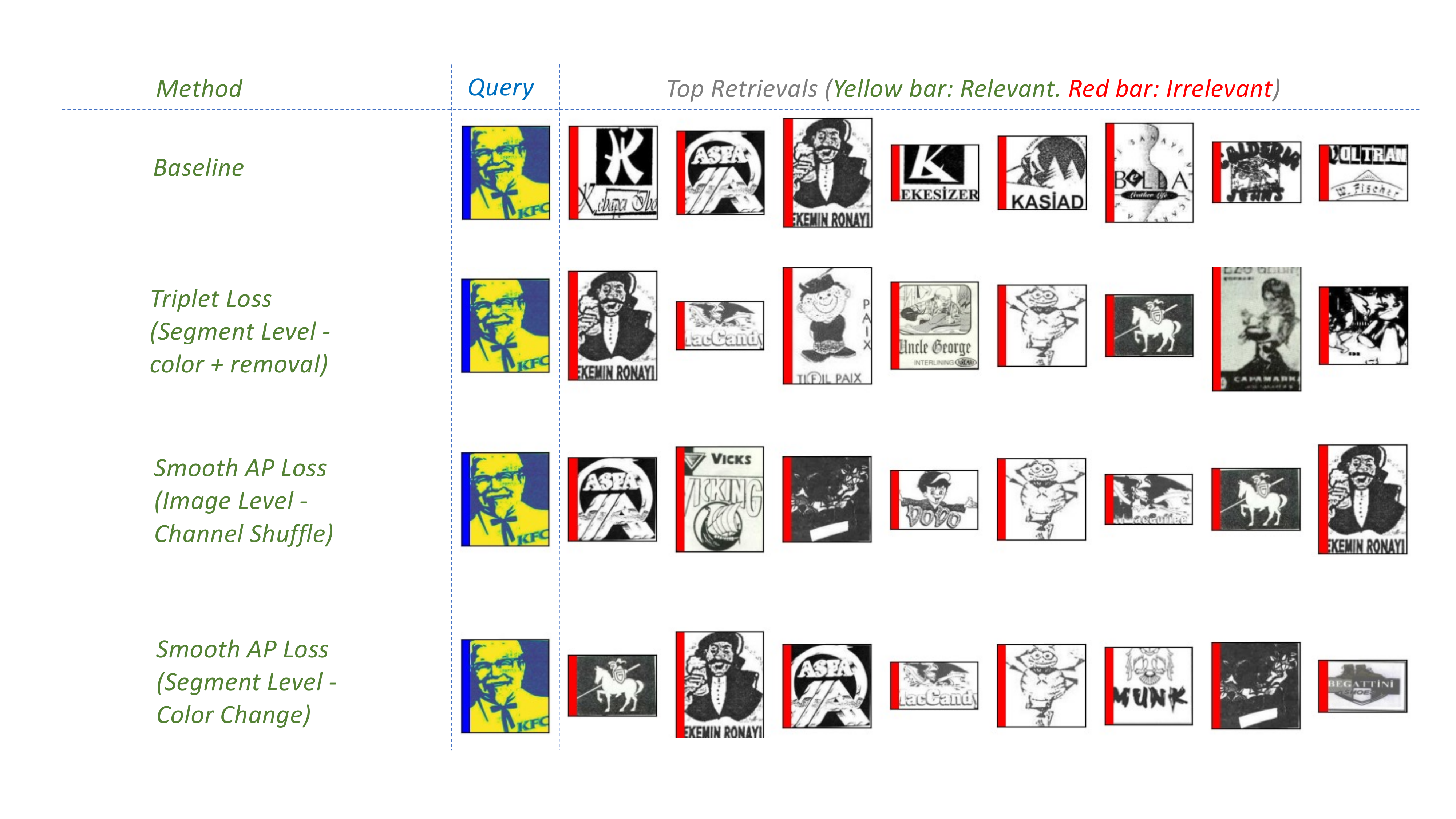}
}}  
\caption{Visual results on the METU dataset for the methods at their best settings. \textbf{(a-b)} Example cases where segment-level augmentation produces better retrieval than image-level augmentation. \textbf{(c)} A negative result for a query with an unusual intensity distribution, which affects all methods adversely. Though we observe that segment-level augmentation is able to retrieve logos with similar color distributions.
\label{fig:visual_results}}
\end{figure*}

Figure \ref{fig:visual_results} displays sample retrieval results for different queries. We see in Figure \ref{fig:visual_results}(a) and (b) that segment-level augmentation is able to provide better retrieval than its competitors. Figure \ref{fig:visual_results}(c) displays a failure case where a query with an unusual color distribution is used. We see that all methods are adversely affected by this; however, segment-level augmentation is able to retrieve logos with similar color distributions.

\subsection{Running-Time Analysis}

Table \ref{tab:RebuttalTimeComplexity} provides a running-time analysis of segment-level and image-level augmentation strategies. We observe that the time spent on segmentation (0.4ms) and Segment Color Change (2.4ms) is comparable to image-level transformation Horizontal Flip (2.8ms). However, Segment Removal and Segment Rotation takes significantly more time. It is important to that our implementation is not optimized for efficiency.

\begin{table}[ht]
    \centering
    \caption{\label{tab:RebuttalTimeComplexity} Time spent on different steps of segment-level and image-level augmentation.}
    \begin{tabular}{@{}lll@{}}
        \toprule
        \multicolumn{1}{c}{\textbf{Method}} &
        \multicolumn{1}{c}{\textbf{Segmentation}} &
        \multicolumn{1}{c}{\textbf{Transformation}} \\ \midrule\midrule
        \multicolumn{1}{c}{Color Jitter} &
        \multicolumn{1}{c}{-} &
        \multicolumn{1}{c}{1.2ms} \\ \midrule
        \multicolumn{1}{c}{R. Horizontal Flip} &
        \multicolumn{1}{c}{-} &
        \multicolumn{1}{c}{2.8ms } \\ \midrule \midrule
        \multicolumn{1}{c}{Segment Color} &
        \multicolumn{1}{c}{0.4ms} &
        \multicolumn{1}{c}{2.4ms} \\ \midrule
        \multicolumn{1}{c}{Segment Removal} &
        \multicolumn{1}{c}{0.4ms} &
        \multicolumn{1}{c}{4.3ms} \\  \midrule
        \multicolumn{1}{c}{Segment Rotation} &
        \multicolumn{1}{c}{0.4ms} &
        \multicolumn{1}{c}{14.0ms} \\
         \bottomrule
        \addlinespace
        \end{tabular}
\end{table}


\subsection{Discussion}

\subsubsection{Comparing Triplet Loss and Smooth-AP Loss}

Our results on the METU and LLD datasets lead to two interesting findings, which we discuss below: 

\begin{itemize}
    \item \textit{Finding 1: Triplet Loss is better in terms of R@1 and R@8 measures whereas Smooth-AP is better in terms of NAR measure.}
    \item \textit{Finding 2: Smooth-AP provides its best NAR performance with Color Change whereas Triplet Loss provides its best with Color Change \& Segment Removal.}
\end{itemize} 
    
Triplet Loss is by definition optimizing or learning a distance metric and considered a surrogate ranking objective. On the other hand, Smooth-AP directly aims to optimize Average Precision, a ranking measure, and therefore, it pertains to a more global ranking objective than Triplet Loss, which works at the level of triplets only. See also Brown et al. \cite{Brown2020eccv} who discussed and contrasted Triplet Loss with Smooth-AP Loss.

The two objectives provides different inductive biases to the learning mechanism and therefore, we see differences in terms of their performances with respect to different performance measures and augmentation strategies. For example, Finding 1 is likely to be because NAR considers all logos whereas R@1 and R8 only consider logos on the top of the ranking which can be easily learned to be ranked at the top using local similarity arrangements as in Triplet Loss. Moreover, Finding 2 occurs because different augmentation strategies incur different ranking among logos and the inductive biases of Triplet Loss and Smooth-AP handle them differently.

We believe that this is an interesting research question that we leave as future work.

\subsubsection{Different Effects and Selection Probabilities for Segment-Augmentations}

The results in Table IV of the main paper show that different augmentations contribute to performance differently and with different selection probabilities. For example, we see that Color Change (with $p=0.5$) gives the best performance in terms of NAR and R@8 whereas Rotation (with $p=0.75$) provides the best performance for R@1 and Removal improves over the baseline only in terms of R@8 measure. 

We attribute these differences to the fact that the different segment-level augmentations incur different biases: Color Change enforces invariance to perturbations in color differences at the segment-level whereas Segment Rotation and Removal encourage invariance to changes to the spatial layout of the shape. 

\subsubsection{Applicability to Other Problems}

We agree that our analysis is limited to logo retrieval. However, the idea of segment-level augmentation is a viable approach for reasoning about similarities at object part levels and require transfer of knowledge at the level of object parts. One good example is reasoning about affordances of objects \cite{zhu2014reasoning,myers2015affordance,myers2014affordance}, where supported functions of object parts can be transferred across objects having similar parts. Another example is reasoning about similarity between shapes that have partial overlap \cite{leonard20162d,latecki2000shape}, where correspondences between parts of shapes need to be calculated. In either example, the specific segment-level augmentation methods may have to be adjusted to the specific problem. For example, performing affine transformations on the segments may be helpful for problems with real-world

\clearpage


\end{document}